\documentclass[Afour,sageh,times]{sagej}

\usepackage{moreverb,url,float,adjustbox,pdflscape,rotating,ragged2e,graphicx,subcaption,xr,rotating,afterpage,booktabs,dblfloatfix}
\usepackage[table,xcdraw]{xcolor}
\usepackage[normalem]{ulem}
\useunder{\uline}{\ul}{}
\usepackage[colorlinks,bookmarksopen,bookmarksnumbered,citecolor=red,urlcolor=red]{hyperref}
\usepackage{tabularx}
\newcolumntype{Y}{>{\centering\arraybackslash}X}
\usepackage{tablefootnote}
\usepackage{subcaption}
\usepackage{graphicx}
\usepackage{amssymb}
\usepackage{multirow,multicol}
\usepackage{makecell}
\usepackage{array}
\usepackage{xspace}
\usepackage{gensymb}
\usepackage{siunitx}
\usepackage{balance}
\usepackage{censor}
\usepackage[flushleft]{threeparttable}
\newcommand\BibTeX{{\rmfamily B\kern-.05em \textsc{i\kern-.025em b}\kern-.08em
T\kern-.1667em\lower.7ex\hbox{E}\kern-.125emX}}

\def\volumeyear{2026}
\newcommand{\ACRONYM}{ACME\xspace} 
\begin{document}


\title{\ACRONYM: A Multi-Cultural, Multi-Embodiment Social-Navigation Dataset}

\author{
Shashank Rao Marpally \affilnum{1*},
Allan Wang \affilnum{2*},
Atharva Ghotavadekar \affilnum{1},
Renato Alexandre Ribeiro \affilnum{2},
Nhat Le \affilnum{3},
Pilar Bachiller-Burgos \affilnum{4},
Pranav Goyal \affilnum{5},
Subham Agrawal \affilnum{6},
Yasuhiro Nitta \affilnum{7},
Howard Ziyu Han \affilnum{8},
Daeun Song \affilnum{3, 10},
Masaki Kuribayashi \affilnum{2},
Kohei Uehara \affilnum{2},
Xiyue Wang \affilnum{2},
Yangzhe Kong \affilnum{3},
Duc M. Nguyen \affilnum{3},
Amirreza Payandeh \affilnum{3},
Gerardo Pérez-González \affilnum{4},
Alejandro Torrejón-Harto \affilnum{4},
Jeeho Ahn \affilnum{5},
Tisha Jain \affilnum{5},
Andrew Stratton \affilnum{5},
Elvin Yang \affilnum{5},
Jorge de Heuvel \affilnum{6},
Nico Ostermann-Myrau \affilnum{6},
Sai Anudeep Sajja \affilnum{6},
Mithilya Raj \affilnum{8},
Daisuke Sato \affilnum{8},
Gaston Rouquette \affilnum{7},
Nikolas Martelaro \affilnum{8},
Maki Sugimoto \affilnum{7},
Hironobu Takagi \affilnum{2},
Chieko Asakawa \affilnum{2},
Maren Bennewitz \affilnum{6},
Aaron Steinfeld \affilnum{8},
Xuesu Xiao \affilnum{3},
Christoforos Mavrogiannis \affilnum{5},
Harold Soh \affilnum{1,9}
}

\affiliation{\affilnum{1} School of Computing, National University of Singapore,
\affilnum{2}Miraikan - The National Museum of Emerging Science and Innovation,
\affilnum{3}George Mason University,
\affilnum{4}Universidad de Extremadura,
\affilnum{5}University of Michigan,
\affilnum{6}University of Bonn,
\affilnum{7}Keio University,
\affilnum{8}Carnegie Mellon University,
\affilnum{9}NUS Smart Systems Institute,
\affilnum{10}Ewha Womans University\\
\affilnum{*}These authors contributed equally to this work.}
\corrauth{Shashank Rao Marpally, School of Computing, National University of Singapore, 13 Computing Dr, Singapore 117417, Email: smarpall@comp.nus.edu.sg}

\begin{abstract}
Understanding how robots and humans move in shared spaces is essential for designing effective social robot navigation policies and predicting human behavior. However, existing datasets often lack the diversity needed to capture differences in culture, geography, and human-robot interaction-factors that strongly shape appropriate social behavior. To address this gap, we introduce \ACRONYM: A Cross-cultural, Multi-Embodiment dataset for social navigation. A large-scale data collection effort across 8 sites in 5 countries, using 7 robot embodiments, \ACRONYM is a large and diverse multi-modal dataset aimed at advancing social navigation research, providing 29.35 hours of onboard robot data and 43.5 hours of overhead pedestrian tracking data. Unlike prior datasets, it focuses on capturing goal-driven social navigation behavior in complex social scenarios with explicit robot-crowd interaction through robot speech. To facilitate learning navigation policies and predicting pedestrian trajectories, \ACRONYM provides 3D and 2D scene features, odometry, interaction information, and human-annotated pedestrian trajectory labels. We make \ACRONYM easy to use by providing both human-readable data for each sensor modality as well as raw binary data. Our qualitative and quantitative analyses show that our dataset captures more challenging scenarios and a broader distribution of pedestrian behavior than previous datasets.
\end{abstract}


\maketitle
\noindent\small\textbf{Preprint notice.}
This arXiv preprint is an exact copy of the manuscript currently under review at
\emph{The International Journal of Robotics Research (IJRR)}.
This version will not be updated during the review process.

\medskip
\begin{figure*}[!ht]
    \includegraphics[width=\linewidth]{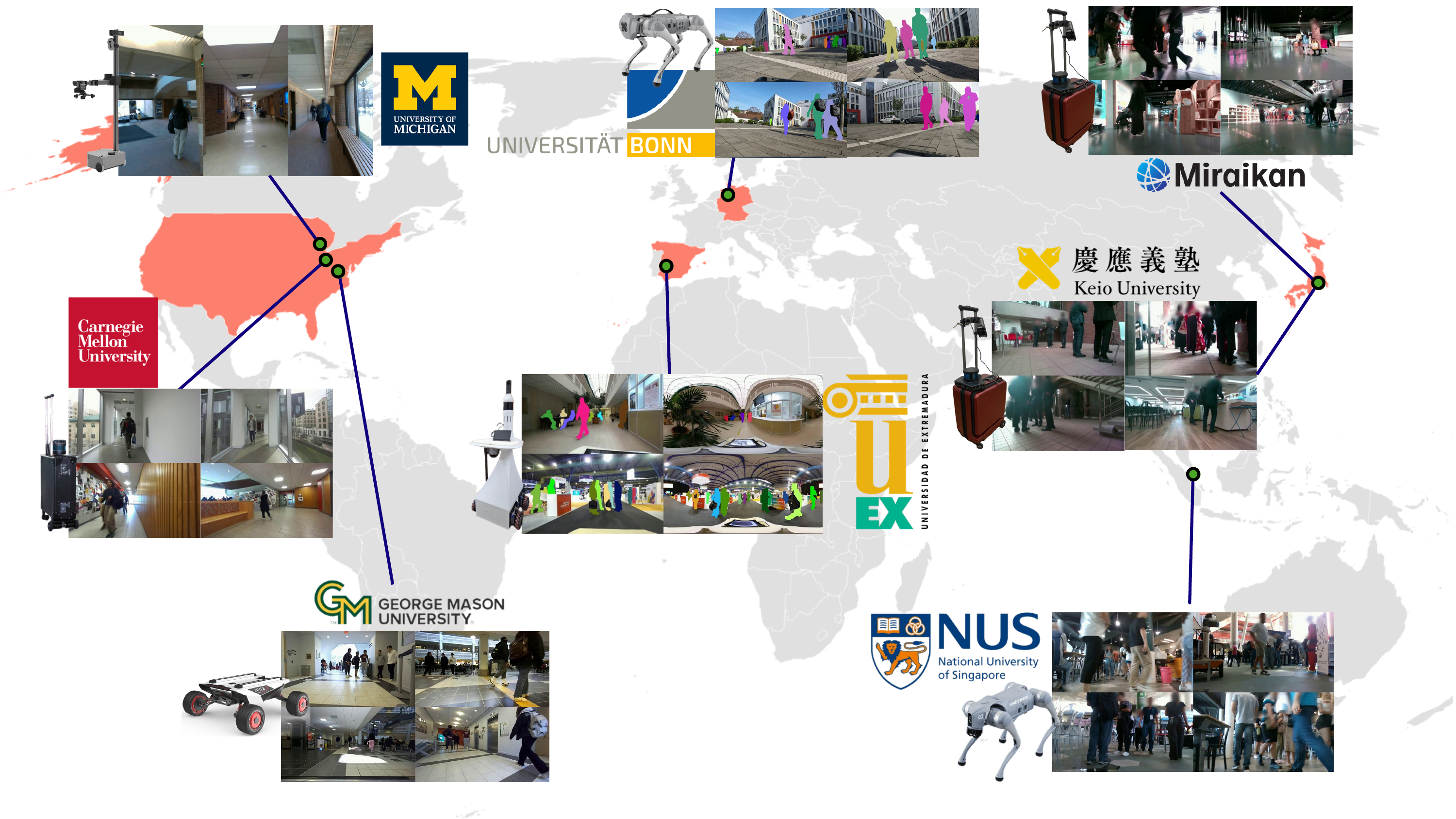}
    \caption{\ACRONYM comprises of data collected by 8 teams across 5 countries on 7 embodiments across diverse crowd scenarios.}
    \label{fig:collaboration_map}
\end{figure*}
\section{Introduction}
Social robot navigation addresses the challenge of enabling autonomous agents to move efficiently and naturally within dynamic human environments. Unlike conventional navigation, which primarily focuses on reaching goals without collisions, social navigation emphasizes compliance with cultural and context-dependent norms that humans implicitly follow in shared spaces~\citep{singamaneni2024survey}. This capability is increasingly critical as robots transition from controlled industrial settings to everyday human-centric applications such as domestic service, food delivery, and security robots. 

Navigating in the presence of humans is also a challenge faced in the autonomous vehicles (AV) domain, and many recent works have focused on achieving safe autonomous navigation~\citep{muhammad2020deep}.
However, the AV setting benefits from widely adopted and well-defined driving rules, driver behavior expectations, and structure in the environment, such as lane markings, traffic lights, and right-of-way laws. Mobile service robots, in contrast, enjoy no such scaffolding, and must reason about and comply with informal, context-sensitive, and often implicit guidelines~\citep{mavrogiannis2023core}.
These include cultural and societal conventions such as how close is “too close” when passing someone, which side of a corridor to stay on, or how to yield in tight spaces.
These rules often vary across cultures and scenarios and are rarely explicitly documented. 

Considerable prior work has sought to formalize the norms governing various aspects of human social navigation, addressing proxemics \citep{kirby2010social}, social signals \citep{takayama2009influences,mead2012probabilistic}, intent communication \& legibility \citep{taylor2022legibility} along with social group dynamics \citep{wang2022group}.
Alongside modeling efforts, complementary work has proposed an evaluation infrastructure for socially compliant navigation, ranging from canonical scenario protocols \citep{pirk2022protocol} to comprehensive guidelines spanning metrics, benchmarks, and datasets \citep{francis2023principles}.

Yet, as critically surveyed by \citep{mavrogiannis2023core}, fundamental challenges persist across motion planning, behavior design, and evaluation. A core reason is that social norms often vary with culture, robot embodiment, and situational context, and multiple norms may apply simultaneously—leading to conflicting constraints that are challenging to identify, model, and reconcile owing to their context dependence. This difficulty in identifying and quantifying these principles and norms of ``correct" social behavior has inspired a perspective shift towards data-driven methods. Specifically, recent works have focused on imitation learning \citep{karnan2022scand,huron,han2025ratatouille}, preference learning \citep{keselman2023pairpref, wang2023navistar, de2023prefdemo}, reinforcement Learning  \citep{liu2023intention, xie2023drl, zhu2026hicrowd}, and lifelong learning \citep{narasimhan2024olivia} approaches for learning social behaviors from expert robot and pedestrian data.
Such methods rely on large, diverse, multi-modal datasets of real-world human–robot navigation for model training, and existing datasets fall short along the axes that matter the most for generalization: diversity in robot embodiments, cultural and geographical contexts, human–robot interaction scenarios, and challenging edge cases \citep{raj2024rethinking}. \\

Closely tied to the challenge of socially compliant navigation is the task of understanding and predicting human motion \citep{rudenko2020trajsurvey}. Social navigation and human motion prediction share the same core problem of understanding human motion. Recently, social navigation models have increasingly integrated trajectory prediction as an important component in their framework \citep{wang2022group, liu2023intention, samavi2025sicnav}. Many datasets have been collected for training and benchmarking human motion prediction models (Table \ref{tab:traj_dset_characteristics}). However, while it is well understood that people from different cultures and environmental contexts exhibit distinct movement patterns and social behaviors, such variance is rarely captured in a single, large-scale dataset. As a result, most predictive models are trained and evaluated on narrow distributions that do not generalize well across culturally or geographically diverse environments, as demonstrated in our benchmark evaluation.
To address these gaps, we propose the \ACRONYM dataset with the following contributions: 
\begin{itemize}
    \item To the best of our knowledge, we release the largest (w.r.t duration) and most diverse (w.r.t geographical location and robot embodiment) human-demonstrated social navigation and pedestrian trajectory prediction dataset from 8 locations across 5 countries and 7 robot embodiments (Fig \ref{fig:collaboration_map}).
    \item We release the largest human-labeled Bird's-Eye View pedestrian trajectory prediction dataset, including camera calibration information to transform these trajectories into the metric space.  
    \item We analyze and compare \ACRONYM to prior social navigation and pedestrian trajectory prediction datasets with respect to scenario complexity and pedestrian trajectory characteristics.
    \item Based on \ACRONYM, we benchmark and compare the performance of SOTA vision navigation and pedestrian trajectory prediction models.
    \item For ease of use, we release human-readable synchronized multi-sensor data, raw ROS2 bags, bird's eye view video, and human-verified pedestrian trajectories. 
\end{itemize}

\section{Related Work}
\subsubsection{Data-Driven Social Navigation}
The landscape of social navigation research is characterized by an extensive variety of approaches \citep{singamaneni2024survey}. Early approaches centered around treating humans as non-reactive obstacles. Social-Force Model \citep{helbing1995social} based approaches like \cite{svenstrup2010trajectory} use proxemics to guide potential-field based planners, thus focusing on spatial social norms. Concurrently, other methods treated the social navigation problem as one of dynamic obstacle avoidance \citep{van2011reciprocal} with constant-velocity models for pedestrians. Reinforcement learning-based approaches utilized such simplified human models to learn navigation in crowds \cite{chen2017decentralized,chen2019crowd,xie2023drl}. More recently, the advent of Vision-Language Models (VLMs) \citep{song2024vlm, xiao2026socialnav, kong2025autospatial, payandeh2025social} and foundational models for navigation \citep{shah2023vint,cheng2024navila} has marked a paradigm shift, towards model-free data-driven navigation policy learning. 
However, the generalization of these modern approaches is fundamentally constrained by the limitations of existing datasets \citep{karnan2022scand,nguyen2023toward,thor} which do not yet capture diverse in-the-wild social-navigation scenes across cultures, geographical locations, and different robot embodiments in a unified fashion.

\subsubsection{Human Motion Prediction}

Beginning with SocialLSTM \citep{alahi2016social}, which standardized trajectory prediction benchmarking, interest in human motion prediction has surged. Beyond early pooling-based \citep{Alahi1,Gupta1} and graph-centric approaches \citep{Social-STGCNN}, recent methods increasingly leverage transformers and diffusion. For instance, transformer architectures unify interaction reasoning and multimodality within a single encoder–decoder pipeline \citep{shi2023tutr}, while diffusion models incorporate social physics or bidirectional consistency to better capture multi-modal futures \citep{chen2024social,li2023bcdiff, panigrahi2023study}. Most recently, a flow-matching-based distillation model has emerged as the new state-of-the-art method \citep{fu2025moflow}. Despite rapid state-of-the-art progress, many studies still benchmark mainly on the ETH/UCY \citep{ETH,UCY} datasets, which are comparatively small-scale and predominantly outdoor, raising robustness and generalization concerns.

Human motion understanding and prediction have also been increasingly relevant to social navigation. It is typically integrated into social navigation models in a two-stage fashion: prediction models first predict pedestrians' future trajectories, and planning modules drive the robot by modeling the predicted trajectories into costs, rewards, or counterfactual information \citep{stratton2026pred2nav, wang2022group, liu2023intention, huron}. A few works have recently emerged that directly leverage human motion prediction models to directly sample trajectories for robots \citep{samavi2025sicnav}, effectively training social navigation in a supervised fashion, yet these models' effectiveness is limited in the real world due to the limited sizes of suitable existing datasets.

\subsubsection{Social Navigation and Human Motion Datasets}
Capturing nuanced variations in pedestrian social behavior, such as those associated with cultural context or the embodiment of nearby navigating agents, requires datasets that span a broad range of interaction settings, agent appearances, and environmental configurations.
Table \ref{tab:comparison_socnav} provides an overview of the distinguishing features between \ACRONYM and prior datasets. Core challenges towards learning social navigation behaviors are addressing the intractability of ``coupled" approaches, the difficulty of designing context-aware social navigation policies, as well as the lack of robust pedestrian behavior prediction models \citep{mavrogiannis2023core}.
Central to addressing these challenges is curating large and diverse datasets to provide a foundation for training models that are predictive, context-aware, and capable of robust generalization to unseen social norms and physical configurations encountered in the real world.
Alongside these characteristics, a notable gap is the absence of datasets that incorporate diverse robot embodiments and explicit crowd interaction, a natural and effective tool that humans use to interactively and cooperatively navigate crowded spaces. 

Trajectory prediction has emerged as an important research direction. However, its focus has traditionally been computer vision related \citep{huang2025vistrajsurvey}. To support robust generalization, it is important to curate data collected in diverse contexts and social norms. To support integration into downstream social navigation models, the dataset should be collected in the real world, its trajectory annotations should be verified instead of relying solely on (potentially error-prone) automated tracking, and most importantly, its pedestrian trajectory coordinates should be grounded in metric coordinates instead of pixel coordinates in the image space. Table \ref{tab:traj_dset_characteristics} provides an overview of the characteristics of prior datasets that support trajectory prediction. Table \ref{tab:bev_stats} further provides dataset size comparisons with prior datasets that satisfy all three key characteristics.

\begin{table*}[!t]
\footnotesize
\setlength{\tabcolsep}{1pt}
\renewcommand{\arraystretch}{1.2}
\caption{Characteristics of all pedestrian datasets that contain pedestrian trajectory annotation}
\label{tab:traj_dset_characteristics}
\centering

\begin{tabularx}{\textwidth}{@{}l*{6}{Y}@{}}
\toprule
& ETH & UCY & Edinburgh & VIRAT & Town Centre & Grand Central \\
& {\scriptsize \cite{ETH}} & {\scriptsize \cite{UCY}} & {\scriptsize \cite{edinburgh}} & {\scriptsize \cite{virat}} & {\scriptsize \cite{towncentre}} & {\scriptsize \cite{grandcentral}} \\
\midrule
\textbf{Location}
& Outdoor & Outdoor & Outdoor & Outdoor & Outdoor & Indoor \\
\textbf{\# Countries}
& 1 & 1 & 1 & 1 & 1 & 1 \\
\textbf{Real-World Data}
& \checkmark & \checkmark & \checkmark & \checkmark & \checkmark & \checkmark \\
\textbf{Verified Trajectories}
& \checkmark & \checkmark &  & \checkmark & \checkmark &  \\
\textbf{Metric Coordinates}
& \checkmark & \checkmark & \checkmark &  & \checkmark &  \\
\end{tabularx}

\vspace{-0.25em}

\begin{tabularx}{\textwidth}{@{}l*{6}{Y}@{}}
\midrule
& CFF & SDD & L-CAS & WildTrack & JRDB & ATC \\
& {\scriptsize \cite{cff}} & {\scriptsize \cite{stanforddrone}} & {\scriptsize \cite{lcas}} & {\scriptsize \cite{chavdarova2018wildtrack}} & {\scriptsize \cite{jrdb}} & {\scriptsize \cite{atc}} \\
\midrule
\textbf{Location}
& Indoor & Outdoor & Indoor & Outdoor & Both & Indoor \\
\textbf{\# Countries}
& 1 & 1 & 1 & 1 & 1 & 1 \\
\textbf{Real-World Data}
& \checkmark & \checkmark & \checkmark & \checkmark & \checkmark & \checkmark \\
\textbf{Verified Trajectories}
&  & \checkmark & \checkmark & \checkmark & \checkmark &  \\
\textbf{Metric Coordinates}
& \checkmark &  &  & \checkmark & \checkmark & \checkmark \\
\end{tabularx}

\vspace{-0.25em}

\begin{tabularx}{\textwidth}{@{}l*{7}{Y}@{}}
\midrule
& TH\"OR & Crowd-Bot & TBD & SiT & TH\"OR-MAGNI & Bi$^3$ & ACME \\
& {\scriptsize \cite{thor}} 
& {\scriptsize \cite{diego2022crowdbot}} 
& {\scriptsize \cite{wang2024tbd}} 
& {\scriptsize \cite{bae2023sit}} 
& {\scriptsize \cite{schreiter2025thormagni}} 
& {\scriptsize \cite{stratton2026bi3dataset}}
& {\scriptsize (Ours)} \\
\midrule
\textbf{Location}
& Indoor & Outdoor & Indoor & Both & Indoor & Indoor & Both \\
\textbf{\# Countries}
& 1 & 1 & 1 & 1 & 1 & 2 & 5 \\
\textbf{Real-World Data}
&  & \checkmark & \checkmark & \checkmark &  &  & \checkmark \\
\textbf{Verified Trajectories}
& \checkmark &  & \checkmark & \checkmark & \checkmark & \checkmark & \checkmark \\
\textbf{Metric Coordinates}
& \checkmark & \checkmark & \checkmark & \checkmark & \checkmark & \checkmark & \checkmark \\
\bottomrule
\end{tabularx}

\end{table*}

\section{Motivation for \ACRONYM}

\ACRONYM is motivated primarily by the observation that contextual factors heavily influence what constitutes acceptable social navigation behavior for robots. \cite{francis2023principles} frames eight principles for evaluating social compliance, including ``contextual appropriateness" -- socially-appropriate robot behavior must consider the various facets of context like culture, environment, task, and interpersonal context. For example, training robots to stay a fixed minimum distance away from all pedestrians might generate favorable behavior in simpler, sparse scenarios while likely resulting in the ``freezing robot" problem \citep{trautman2010unfreezing,trautman2015robot} in overly crowded or geometrically constrained scenarios. Similarly, robots trained to always move to one side of a hallway may appear socially compliant in some countries and noncompliant in others. Additionally, within the same geographical location, different geometrical layouts of navigation environments, crowd densities, and robot embodiments necessitate adaptive navigation strategies--what is perceived as safe and appropriate in a spacious corridor with sparse pedestrian traffic for a vacuum cleaning robot may be interpreted as inefficient or even obstructive in a narrow, densely populated walkway for a robot dog. These nuances highlight the need for datasets that capture multi-cultural and multi-embodiment data in diverse environments, enabling the development of context- and embodiment-aware navigation policies. 
This motivation also informs our data-collection strategy: rather than treating diversity as an incidental property of the dataset, we explicitly align \ACRONYM with prior recommendations for social-navigation dataset curation \citep{francis2023principles}, including broad but resource-aware coverage, well-sampled scenarios, real robot behavior, diverse robot platforms, and systematic annotations. We now highlight the motivations behind the unique features of \ACRONYM.

\subsection{Capturing Cross-Cultural Social Norms}  
Cultural differences, personal preferences, and environmental factors heavily influence pedestrian motion characteristics. For example, a large-scale study (\cite{sorokowska2017preferred}) found significant variance in the preferred social, personal, and intimate distance for interaction across country, age, and gender. It is also well known that the direction of traffic flow varies across countries. For example, in the United States, Spain, and Germany, people prefer to walk on the right side of a walkway, while in Singapore, Japan, and India, the left side is preferred. Intending to train robots and human motion prediction models to account for such socio-cultural contexts, we curate data collected across 5 countries in 3 continents (8 different data collection teams; 7 of the 8 locations are university campuses, while Miraikan is a science museum). This distributed data collection strategy helps capture differences in navigation and communication styles of both pedestrians and the robot, thus facilitating learning of culturally appropriate behaviors.  

\subsection{Diversity in Robot Morphology}
Social robotics studies have extensively documented the effect of robot morphology on the perception, likability, and nature of interaction of the robot with a user \citep{bradwell2021morphology,haring2016people}. For example, the form--function attribution bias (FFAB) describes how users may infer a robot's capabilities, intelligence, or social competence from its physical appearance rather than from its actual functional abilities \citep{haring2018ffab}. This suggests that robot embodiment can shape human expectations even before any interaction occurs. Different robot morphologies also differ in navigational affordances and offer an opportunity to capture the variance in the behaviors of pedestrians towards different robots operating in the wild. Our dataset utilized 5 visually distinct robots with different navigation styles and sensor setups to gather data (Table \ref{tab:data_overview_table}): namely, 2 quadrupeds (NUS, UBonn), 2 suitcase-style differential drive robots physically guided by human operators (CMU, Miraikan, Keio), 1 rover-style differential drive robot (GMU), and 2 mobile service robots (UMich, UEx). 
Despite this variance in setups, we maintain consistency across these subsets of our data by maintaining a minimum set of sensor data modalities and common teleoperation and data-recording guidelines. 

\subsection{Curating Local Social Behaviors}
We distinguish \ACRONYM from other valuable datasets \citep{karnan2022scand,nguyen2023toward} by focusing on the quality of data, and targeting our data collection effort towards capturing challenging social navigation scenarios in-the-wild. Particularly, we ensure that in each trajectory the robot encounters at least one human, while preferring scenarios with varying levels of crowds and challenging location configurations like blind corners, intersections, and narrow corridors. Our data collection teams sought out locations based on the guidelines of \cite{francis2023principles} and timings that coincided to provide maximum crowd encounters with the robot. In order to curate useful interactions and scenarios, we additionally augment the teleoperation procedure with guidelines and tools to facilitate the collection of goal-oriented, high-quality robot trajectory data. 

\subsection{Social Navigation with Robot Speech}
Pedestrians often use verbal and non-verbal cues when asking for directions, alerting other social actors of their presence, or apologizing when causing disruptions to other pedestrians. Learning such communication behavior is desirable for robots to adapt to various levels of crowds and navigate in a trustworthy and safe manner \citep{che2020efficient,kannan2021external,hart2020using}. However, there is a lack of real-world datasets focused on learning such communication policies from expert behavior. We thus include a set of phrases that the robot can utter and record when and what was spoken by the robot. This places additional constraints on the teleoperator and robot platform, thus only a subset of \ACRONYM includes robot speech data, namely, the NUS, UEx, Miraikan, and Keio subsets. Paired with scene information obtained from LiDAR and FPV RGB, utterance data opens avenues for teaching robots to communicate effectively to proactively influence the crowd behavior.

\subsection{Understanding Context Conditioned Human Behavior}
Besides the data from robots manually operated by humans, social navigation has increasingly integrated trajectory prediction models as part of the broader navigation framework. Some works predict trajectories first, and then plan robot controls by assigning costs to the predicted trajectories \citep{wang2022group, poddar2023crowd}. Other works integrate trajectory prediction outputs as additional rewards or context \citep{liu2023intention, huron}. Explicitly, more accurate trajectory prediction may lead to better downstream navigation performance. Implicitly, studying how humans move in the same environment also offers valuable insights into how robots should move. Although robot data are of higher quality and realism, containing goal annotations, and are not anthropomorphic~\citep{epley2007anthropomorphism}, human behavior data are much richer and greater in quantity. 

In the \ACRONYM dataset, we label human trajectories from the overhead Bird's-Eye View (BEV) cameras and transform the pixel coordinates into metric coordinates via calibrated sensor information. By doing so, we obtain data on how humans navigate given their current surrounding context. Because multiple humans are often present in our dataset sessions, and their combined trajectories cover wider areas in the dataset environments, human trajectory data captures ground truth human navigation behavior more comprehensively, given their greater quantity and diversity. Recent works have emerged, such as SICNav-Diffusion \citep{samavi2025sicnav}, that directly leverage trajectory prediction as a trajectory sampler. Although their effectiveness in the real world is capped by the limited size and variety of existing datasets, they revealed that a comprehensive human motion dataset may enable additional supervised research approaches to social navigation.

\section{The \ACRONYM Dataset}
In this section, we describe the methodology and nature of the data captured in \ACRONYM \footnote{All software developed in this section will be open-sourced.}.
An overview of the data duration, hardware, sensors, and locations for each data collection team is summarized in Table \ref{tab:data_overview_table}.
\begin{table*}[t!]
    \centering
     \caption{Overview of data characteristics and hardware setup for each subset of \ACRONYM.}
    \renewcommand{\arraystretch}{1.2} %
    {\footnotesize
    \begin{tabularx}{\textwidth}
    {X|XXXXXXXX}
        \toprule %
        Information \ Subset&
        National University of Singapore (\textbf{NUS})&
        Museum of Emerging Science and Innovation (\textbf{Miraikan})&
        Carnegie Mellon University (\textbf{CMU})&
        Keio University (\textbf{Keio})&
        University of Michigan (\textbf{UMich})&
        University of Extremadura (\textbf{UEx})&
        George Mason University (\textbf{GMU})&
        University of Bonn (\textbf{UBonn})\\
        \midrule%

        Robot Platform&
        Unitree GO2\footnotemark[1]&
        Suitcase robot\footnotemark[2]&
        Suitcase robot\footnotemark[2]&
        Suitcase robot\footnotemark[2]&
        Stretch 2\footnotemark[3]&
        Shadow Robot\footnotemark[4]&
        Scout Mini\footnotemark[5]&
        Unitree GO1\footnotemark[6] \\
     
        Sensors& 
        Realsense D435i, Hesai XT-16 &%
        Realsense D435i, Velodyne VLP-16 &%
        Zed 2, Velodyne VLP-16&
        Realsense D435i, Velodyne VLP-16&
        Realsense D435i, Velodyne VLP-16&
        Zed 2i, Ricoh Theta z1, Robosense BPearl \& Helios&
        Zed 2, Velodyne VLP-16&
        GoPro, Velodyne VLP-16\\
        
        Pose Estimate&
        LiDAR-Odometry&
        LiDAR Localization&
        LiDAR Localization&
        LiDAR Localization&
        Wheel Odometry&
        Visual-Inertial Odometry, LiDAR localization&
        Wheel Odometry&
        LiDAR-Odometry  \\
        
        BEV Camera&
        GoPro Hero 9&    %
        GoPro Hero 12&   %
        GoPro Hero 10&   %
        GoPro Hero 12&   %
        GoPro Hero 13&   %
        - & 
        Logitech C920&
        GoPro Hero 12 \\
        
        Map-based Localization & No & Yes  & No & Yes & No & No & No & No \\
        
        Data Duration (hrs) & 6.54 & 7.53 & 1.56  & 1.35 & 3.2 & 3.29 & 3.45  & 2.43 \\

        BEV Data Duration (hrs) & 10.01 & 12.39 & 2.65  & 3.84 & 5.10 & - & 2.30  & 7.27 \\

        BEV \# Trajectories & 31848 & 21322 & 1987  & 5095 & 3728 & - & 3186  & 4892 \\
        \bottomrule%
        
    \end{tabularx}
    }
   
    \label{tab:data_overview_table}
    \vspace*{-1.0\baselineskip}
\end{table*}

\subsection{Hardware and Software Setup}
To facilitate consistency and quality across the dataset while accommodating differences in robot platforms and sensing modalities, we adopt a set of common guidelines for data collection and curation.
We ensure that all trajectories comprise of locally accurate robot poses (via Odometry), Egocentric RGB Camera feeds (for semantic scene understanding), 3D LiDAR point clouds (for spatial reasoning), extrinsic and intrinsic sensor calibration (for sensor-fusion). A subset of the dataset also includes Robot Speech (timing and utterance) for crowd interaction. Onboard sensor data was collected in ROS2 \citep{macenski2022robot} bag format, then post-processed for anonymization and downsampled to provide human-readable data. Table \ref{tab:data_overview_table} describes the hardware for data collection in each location. Additionally, several data-collection teams also collected additional sensor information, for example, accurate robot localization with pre-built maps (Miraikan, CMU, and Keio University teams), and egocentric 360 RGB images (GMU and UEx teams). Each subset of data is also accompanied by sensor calibration for robot-onboard sensors performed with off-the-shelf techniques, allowing for multi-modal scene understanding. 

In addition to robot data, NUS, Miraikan, CMU, Keio, U-Mich, GMU, and UBonn subsets have overhead BEV cameras set up overlooking the data collection areas. The BEV cameras collected human trajectory data from a less constrained perspective. Since humans in the scene are at least partially observable from the overhead cameras, automated tracking methods were used to assist human trajectory annotations~\citep{wang2024tbd}.

\begin{figure}
    \centering
    \includegraphics[width=1.0\linewidth]{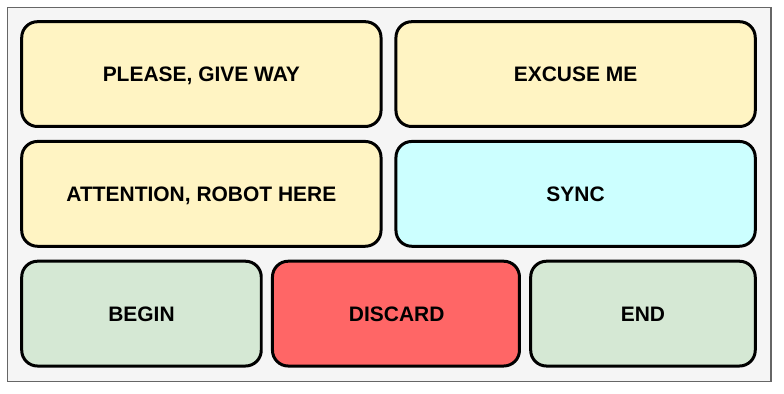}
    \caption{Teleoperation interface used to collect data in NUS where the teleoperator can relay robot speech commands and mark start and end of trajectory segments during data collection.}
    \label{fig:tele_app}
\end{figure}    
\begin{figure}
    \includegraphics[width=1.0\linewidth]{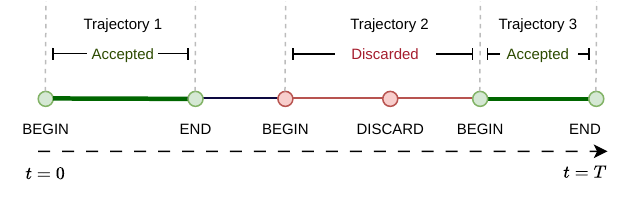}
    \caption{Data filtering with Event Markings}
    \label{fig:data_splitting}
    \vspace*{-1.0\baselineskip}
\end{figure}
\subsubsection{Teleoperation Guidelines} 
\begin{figure}[t]
     \centering
     \begin{subfigure}[b]{0.5\textwidth}
         \centering
         \includegraphics[width=\textwidth]{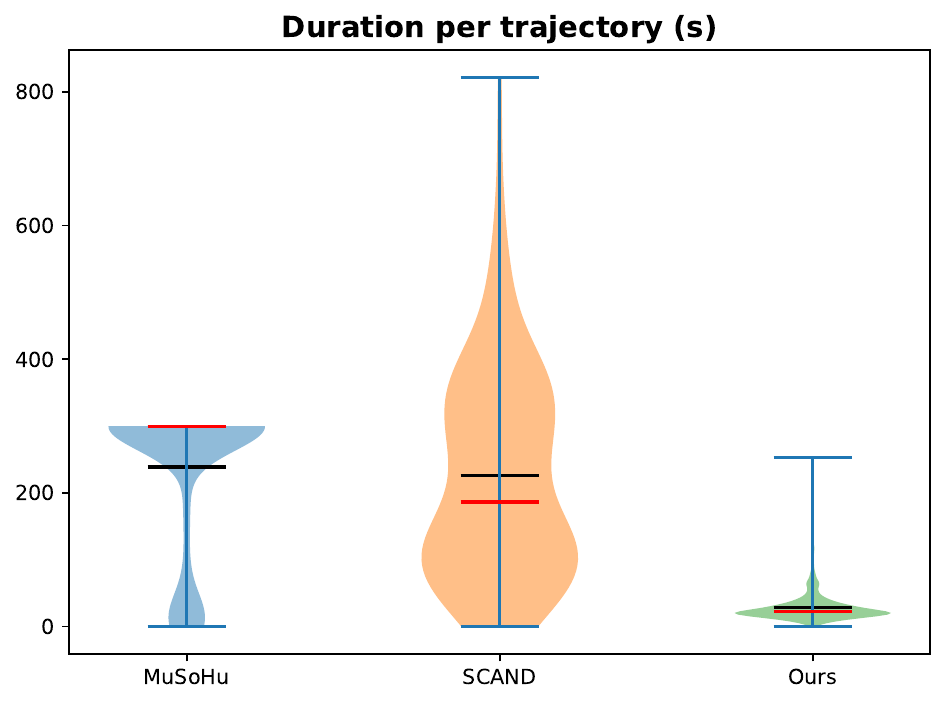}
         \caption{Distribution of individual trajectory durations.}
     \end{subfigure}
     \hfill
     \begin{subfigure}[b]{0.5\textwidth}
         \centering
         \includegraphics[width=\textwidth]{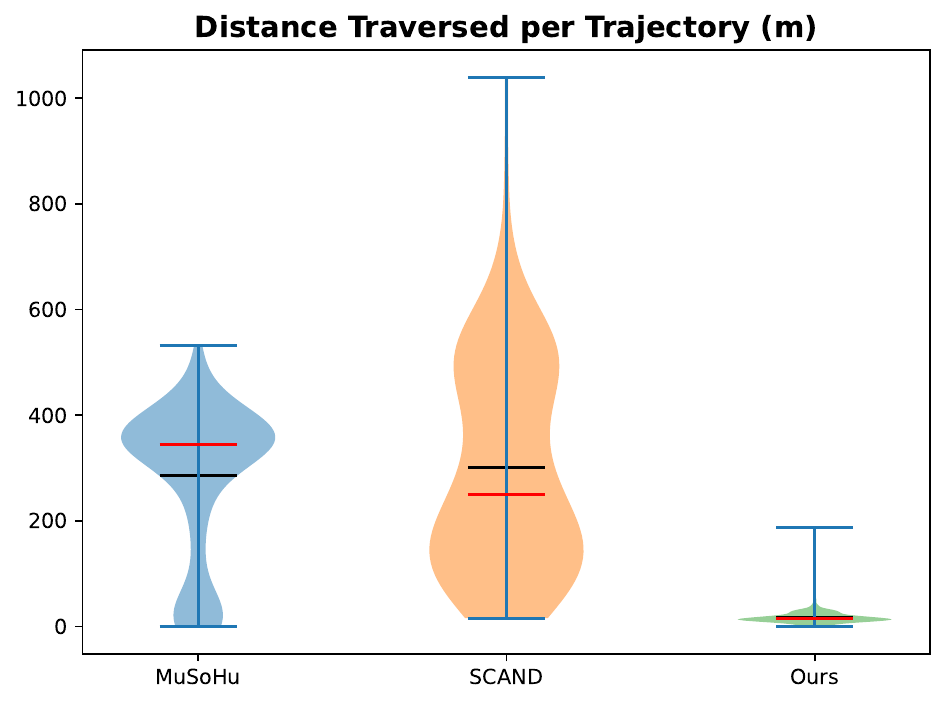}
         \caption{Distribution of distance traversed in individual trajectories.}
     \end{subfigure}
     \caption{Compared to prior Social Navigation Datasets that collect unconstrained long trajectories, \ACRONYM trajectories are more localized and goal-oriented.}
     \label{fig:dist_dur_stats}     
\end{figure}
Recognizing the limitations of prior data collection methodologies, we established the following teleoperator guidelines for maximizing the utility of the dataset and capturing natural human-robot-interactions in the wild:
\begin{itemize}
    \item \textbf{Stay as far away from the robot as safely possible}: To emulate ``Wizard of Oz" conditions, teleoperators are advised to stay away from the robot and provide pedestrians the illusion that the robot is navigating autonomously. (Note: this does not apply to the human-operated CMU, Miraikan, and Keio suitcase robots)
    \item \textbf{Use speech when logical}: We use a restricted set of speech commands relayed through a speaker mounted on the robot, which can be triggered through the teleoperation interface (or via the teleoperator verbally in case of the suitcase robots) to standardize behaviors across the dataset. The set of verbal commands is Excuse Me", ``Attention, Robot Here", ``Please Give Way". An example of the interface (used by the NUS team) is shown in Fig \ref{fig:tele_app}. 
    \item \textbf{Fix a navigation goal apriori to starting data collection and record ``Begin", ``Discard", ``End" signals for marking trajectory segments}: To ensure each trajectory describes a logically appropriate path that we expect a robot to follow to reach a predefined navigation goal, teleoperators are asked to fix (implicitly or explicitly via a marking in the environment) a location target for navigation before collecting a trajectory. Each teleoperator also marks events on the respective teleoperation interface that correspond to the start and end of a trajectory, as well as situations where the trajectory must be discarded (for example, if the teleoperator is approached by a pedestrian for conversation or the robot behaves erratically due to hardware issues, etc.). The final dataset is obtained by filtering the raw data with the scheme shown in Fig \ref{fig:data_splitting}.
\end{itemize}
The difference in data-collection strategy between \ACRONYM and prior social navigation datasets like SCAND \citep{karnan2022scand} and MuSoHu \citep{nguyen2023toward} is immediately apparent from the distribution of individual trajectory duration and traversal distance (Fig \ref{fig:dist_dur_stats}): while SCAND and MuSoHu are largely in-the-wild datasets where the data-collection agents wander through locations for long durations and distances, \ACRONYM focuses on capturing short goal-directed scenarios. 
\footnotetext[1]{Unitree GO2 EDU - \url{https://www.unitree.com/go2}} 
\footnotetext[2] {Cabot robot - \cite{cabot_latest} \url{https://www.miraikan.jst.go.jp/en/lab/AIsuitcase/}}
\footnotetext[3] {Shadow robot \cite{shadow_robot_2024} - \url{https://robolab.unex.es/en/robots/shadow/} }
\footnotetext[4]{Stretch - \url{https://hello-robot.com/stretch-2-product}}
\footnotetext[5] {AgileX Scout mini - \url{https://global.agilex.ai/products/scout-mini}} 
\footnotetext[6] {Unitree GO1 - \url{https://www.unitree.com/go1}}
\subsubsection{Transformation from BEV Image Plane to Ground Plane} 
\begin{figure}
        \includegraphics[width=1\linewidth]{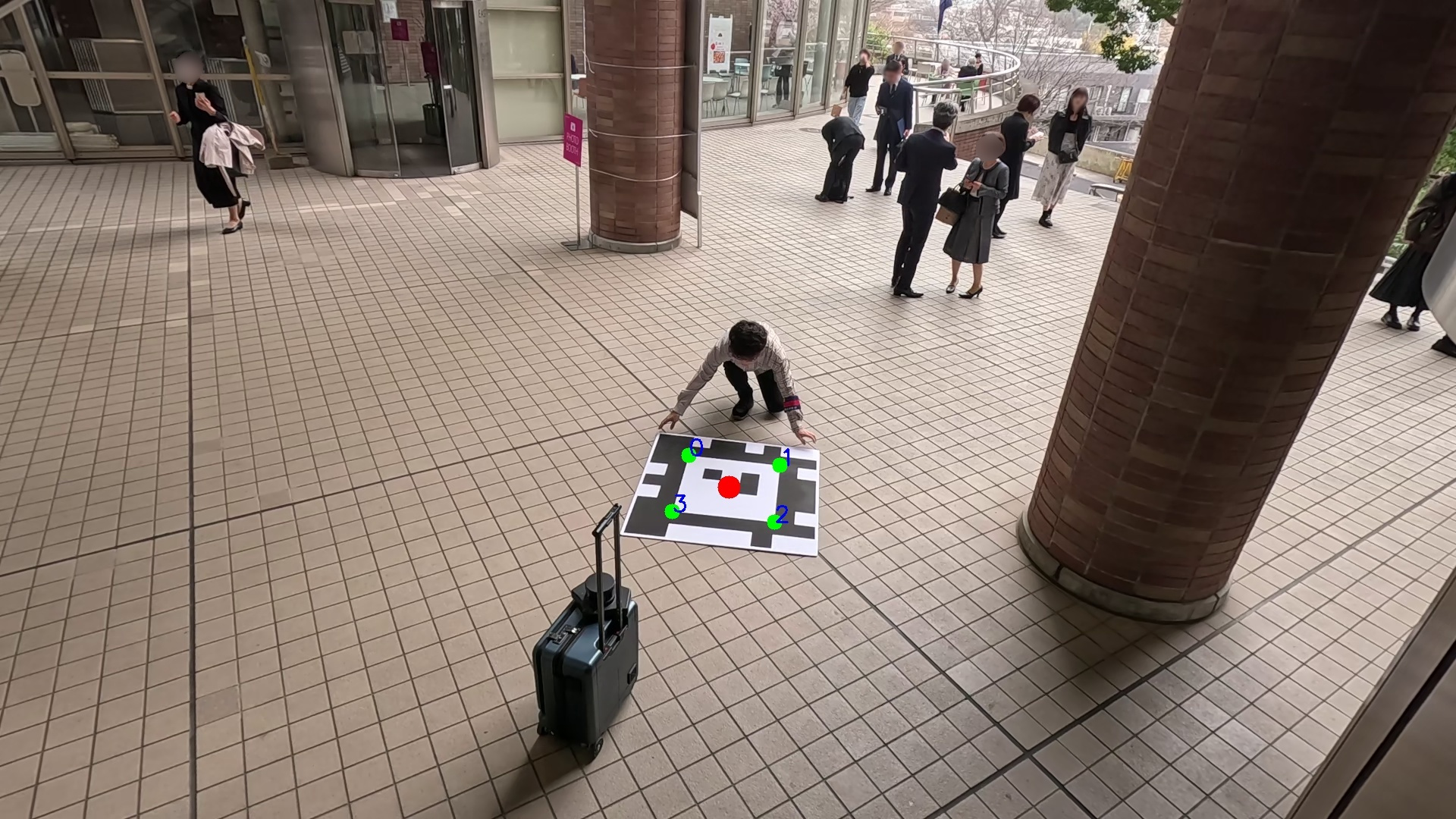}
        \caption{Position Synchronization with April Tags. The tag is visible to both the robot (suitcase) and the BEV camera.}
    \label{fig:pos_sync_keio}
\end{figure}

In order to obtain ground plane metric coordinates after the human trajectories were annotated in the BEV camera frames, 2D to 2D homography transformations were estimated. For Miraikan, Keio, GMU, and UBonn, this is done via a poster-sized AprilTag~\citep{olson2011apriltag} placed on the ground plane briefly during data collection (Fig. \ref{fig:pos_sync_keio}). Given that the tags were laid flat on the ground plane, homographies can be easily obtained via tag detection and pose estimation. While we acknowledge the inherent inaccuracies of tag-based homography estimation, these methods represent a pragmatic compromise between the complexity of full SLAM-based map building in dynamic environments and having no spatial calibration at all. For CMU and U-Mich, ground truth stationary keypoints (e.g., tile intersections) were selected and measured at each location. The keypoints' corresponding pixel locations were obtained in the BEV images, and the resulting point correspondences were used to estimate homographies. For NUS, the poster-sized AprilTags were held upright and can be seen by both the robot camera and the BEV cameras. However, simply assuming the tags to be perfectly vertical resulted in homography estimation errors. Instead, we first obtained the robot's pose with respect to the ground plane-based world frame at the time when it sees the upright tag. We then estimated the BEV cameras' poses in the world frame via tag detection at the synchronized time from both the robot and the BEV cameras. Homographies were then extracted from the cameras' world frame poses.

\subsection{Post Processing}

\subsubsection{Data Anonymization} 
\begin{figure*}
    \centering
    \begin{subfigure}{0.32\textwidth}
        \includegraphics[width=\linewidth]{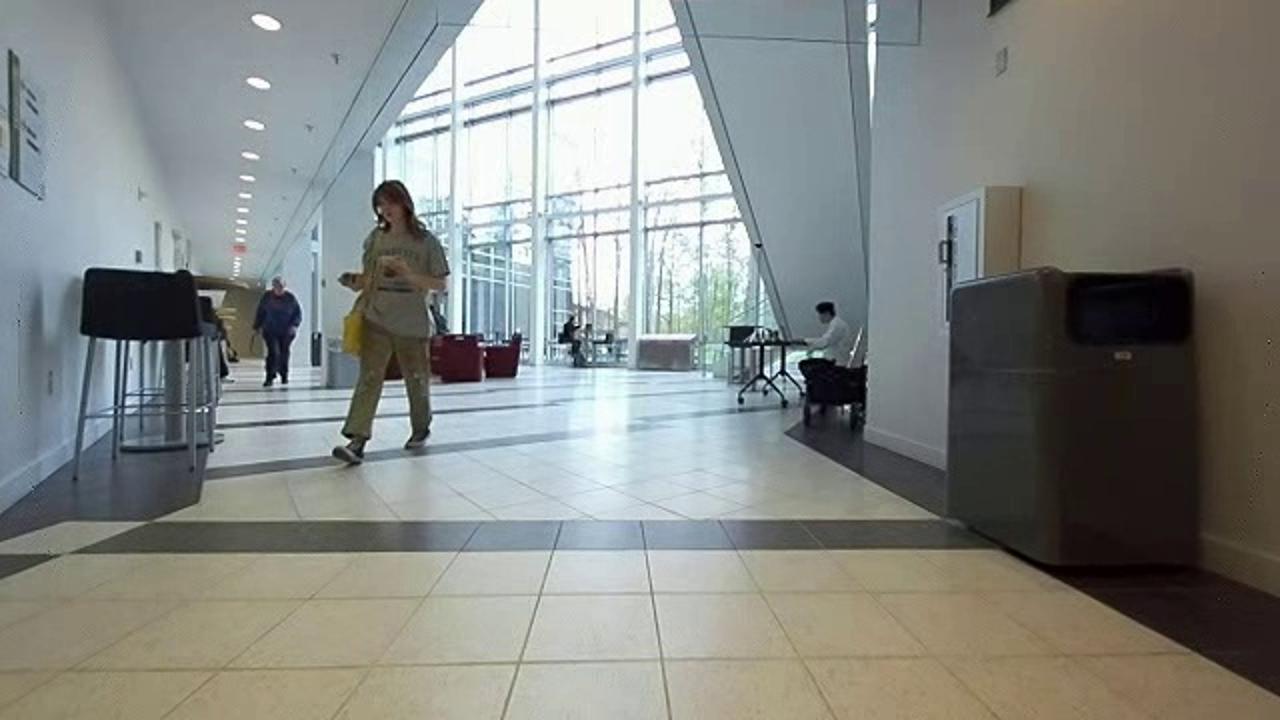}
        \caption{No Anonymization (GMU)}
    \end{subfigure}
    \hfill
    \begin{subfigure}{0.32\textwidth}
        \includegraphics[width=\linewidth]{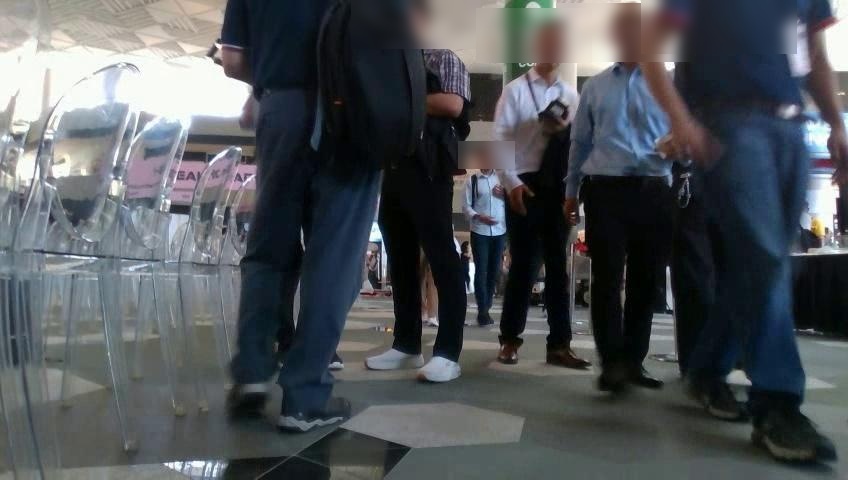}
        \caption{Face Blurring (NUS)}
    \end{subfigure}
    \hfill
    \begin{subfigure}{0.32\textwidth}
        \includegraphics[width=\linewidth]{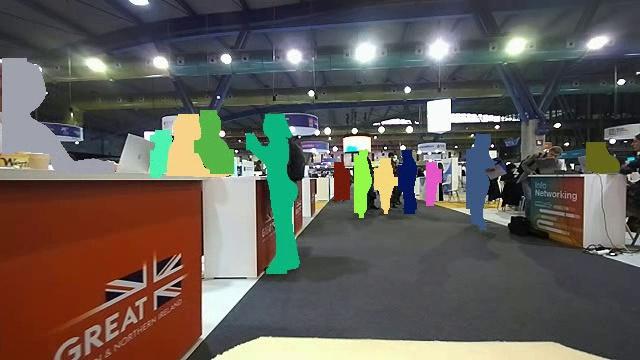}
        \caption{Full Body Segmentation (UEx)}
    \end{subfigure}
    \caption{Different types of Data Anonymization across \ACRONYM specified by the IRB requirements for each institution. In order to retain the lost information, pedestrian tracking information used to anonymize the data (bounding box, segmentation mask, and keypoints) is provided.}
    \label{fig:anonymization_samples}
    \vspace*{-1.0\baselineskip}
\end{figure*}

Due to our data collection teams being spread out around the globe, each team's data had to abide by the respective Institutional Review Board's privacy guidelines for open-sourcing data collected by onboard and BEV sensors. Notably, there were different levels of anonymization required for the humans captured in RGB data across different institutions. These include (in increasing order of information loss) as shown in Fig \ref{fig:anonymization_samples}: 
\begin{itemize}
    \item No anonymization (GMU, CMU)
    \item Face blurring (NUS, UMich, Keio)
    \item Full body segmentation (UEx, Miraikan, UBonn)
\end{itemize}

Although anonymization is essential to preserve privacy, it leads to a loss of identifying features, making it difficult to use the data with off-the-shelf vision models. To compensate for this information loss, we release bounding box, segmentation mask, and pose estimation (keypoint) detections for the anonymized pedestrians using Yolov8 (\cite{Jocher_Ultralytics_YOLO_2023}) and ByteTrack\footnote{ByteTrack Github - \url{https://github.com/FoundationVision/ByteTrack}}~\citep{zhang2022bytetrack}.
\begin{figure}[ht]
    \centering
    \includegraphics[width=1.0\linewidth]{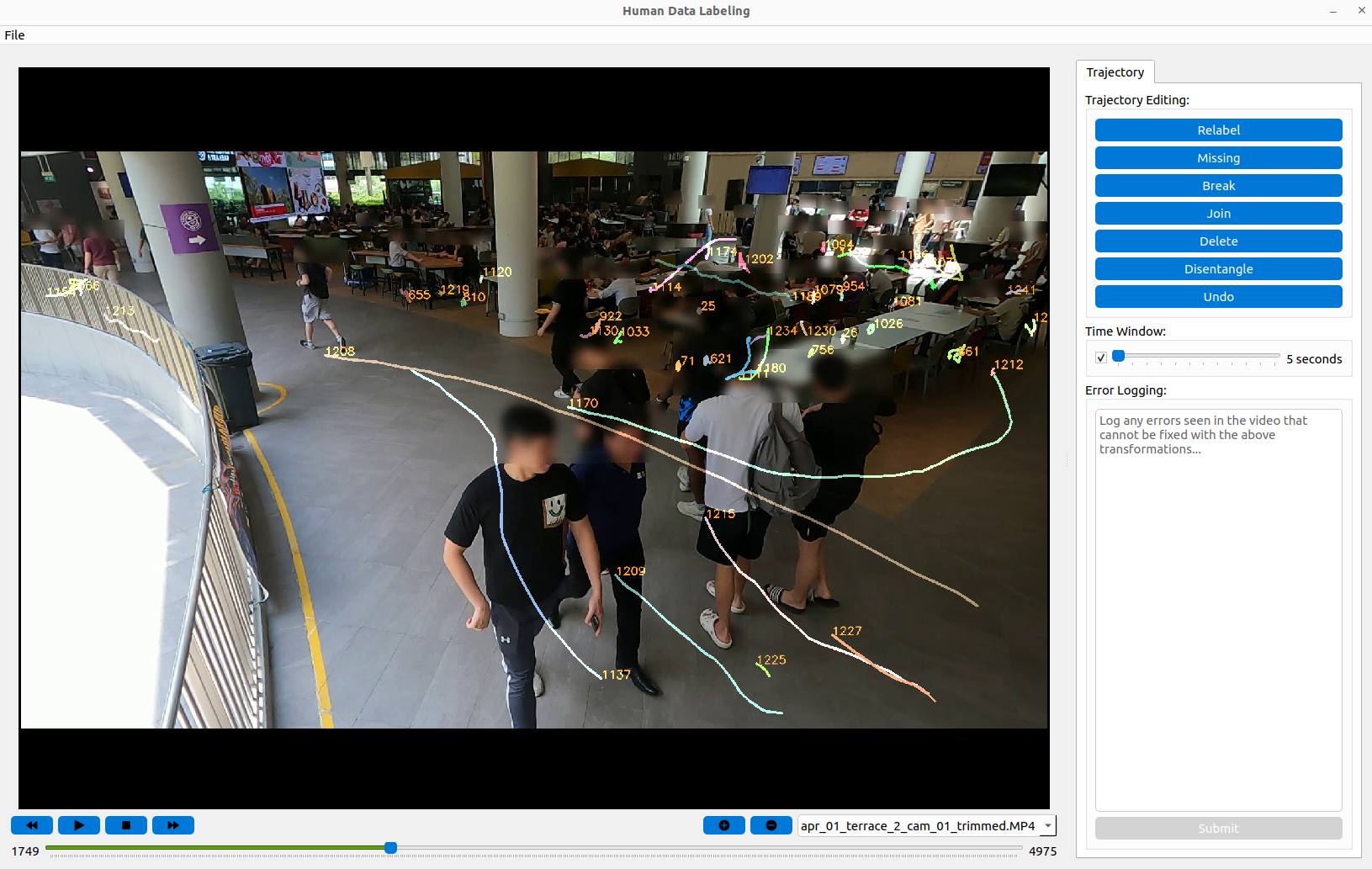}
    \caption{Application interface for human verification process. It contains a media player and multiple options to correct tracking errors manually.}
    \label{fig:labelling-tool}
\end{figure}
\subsubsection{Pedestrian Trajectory Tracking and Correction in BEV} - Prior work on trajectory datasets widely employs semi-automatic annotations, particularly on pedestrian datasets---many of these datasets are relatively small in scale ~\citep{pellegrini2009never, chavdarova2018wildtrack, martin2023jrdb} or lack annotations of the pedestrian surroundings ~\citep{karnan2022socially, granados2022pedestrian}---however, recent efforts have begun addressing these gaps ~\citep{wang2024tbd}. 

Similar to~\cite{wang2024tbd}, we first performed automated tracking of pedestrians from the BEV cameras using ByteTrack.
To ensure the annotation quality of our large-scale dataset, human verification of the tracked trajectories was performed at 10Hz. Based on the features of existing tools~\cite{wang2024tbd} that streamline the human verification process, we designed our own annotation tool to simplify frame-by-frame human annotation.

Our human verification tool (Fig \ref{fig:labelling-tool}) includes a media player that allows users to view videos with automatically generated trajectories by ByteTrack. If an error is detected by the human annotator users, users can correct it using the available editing options:

\begin{itemize}
    \item \textbf{Relabel}: Used to redraw a trajectory from a specific frame when ByteTrack assigns incorrect or imprecise paths.
    \item \textbf{Missing}: Used to manually add a trajectory when ByteTrack fails to detect a pedestrian.
    \item \textbf{Break}: Used when ByteTrack incorrectly assigns the same trajectory to two different pedestrians.
    \item \textbf{Join}: Used to merge two trajectories into a single continuous path.
    \item \textbf{Delete}: Used to remove a trajectory from the current frame onward.
    \item \textbf{Disentangle}: Used to swap back two overlapping trajectories after a specific frame to correct misassigned identities.
    \item \textbf{Undo}: Reverts the most recent trajectory modification, working backwards through the session's history.
\end{itemize}

\begin{figure*}
    \centering
    \includegraphics[width=1\linewidth]{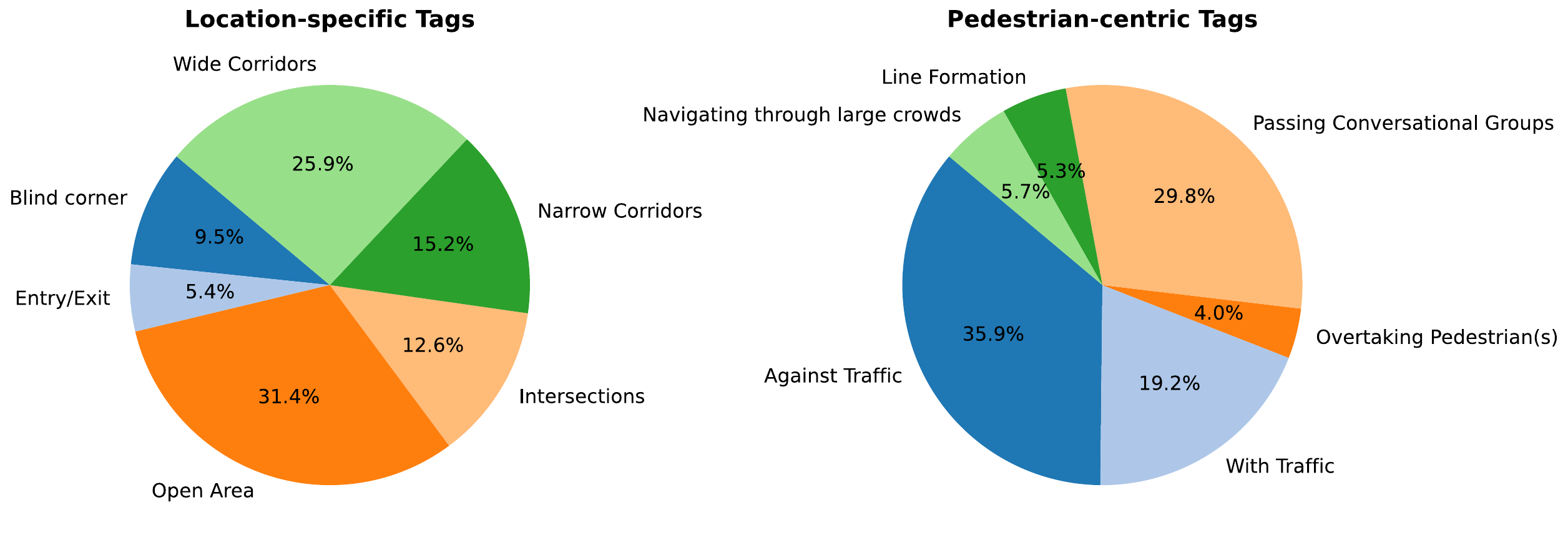}
    \caption{Distribution of Scenarios captured by \ACRONYM. Scenarios dependent on the type of location where the robot and pedestrians navigate are \textit{Location-specific scenario tags} while those characterized by the type of relative motion between the pedestrians and the robot are \textit{Pedestrian-centric} Scenario Tags
    }
    \label{fig:scenario_pie}
    \vspace*{-1.0\baselineskip}
\end{figure*}

\subsubsection{Scenario Tagging of Robot Data}
To enable filtering the dataset for specific scenarios, we additionally developed another annotation tool (a modification of VIA \citep{dutta2019via}) that allows annotators to manually tag robot trajectories for scenarios that occur within them. We identify scenarios based on previous works (\cite{francis2023principles,karnan2022socially,nguyen2023toward}) and include both location-based and pedestrian-based scenario tags. Fig.~\ref{fig:scenario_pie} shows the distribution of scenarios in \ACRONYM, and the description of each tag is provided in Table \ref{tab:scenario_tags_nus}.
\begin{table}
\caption{Tags used to characterize each trajectory in \ACRONYM} 

    \raggedright
    \captionsetup{justification=justified, singlelinecheck=false}
    \begin{tabular}{
    >{\raggedright\arraybackslash}p{0.25\linewidth}
    >{\raggedright\arraybackslash}p{0.4\linewidth}
    >{\centering\arraybackslash}p{0.2\linewidth}}
    \toprule
         Scenario Tag &Tag Description& \# Tags\\
         \midrule %
         Narrow Corridors &Corridors that are ~1-2 pedestrian wide& 864\\\hline
         Wide Corridors &Corridors that are ~4-5 pedestrian wide& 1468\\\hline
         Intersections &Intersection of multiple corridors/walkways& 713\\\hline
         Against Traffic &Against oncoming pedestrian traffic& 2234\\\hline
         With Traffic & Alongside pedestrian traffic& 1196\\\hline
         Passing Conversational Groups &Past a group of 2 or more people that are talking amongst themselves& 1854\\\hline
         Blind corner &Past a corner which the robot cannot see past& 536\\\hline
         Open Area & Areas with minimal spatial constraints& 1781\\\hline
         Navigating through large crowds &Through large unstructured crowds& 352\\\hline
 Entry/Exit &Through doorways/Elevators&306\\\hline
 Line Formation (like queues) & Across people waiting in a line&328\\\hline
 Overtaking Pedestrian(s) &Overtaking a person or groups of people &251\\
    \bottomrule%
    \end{tabular}
\label{tab:scenario_tags_nus}
\end{table}

\subsubsection{Semantic Tagging of BEV Scenes}
\label{sec:bev_semantic_tag}
\begin{figure}
    \centering
    \includegraphics[width=1\linewidth]{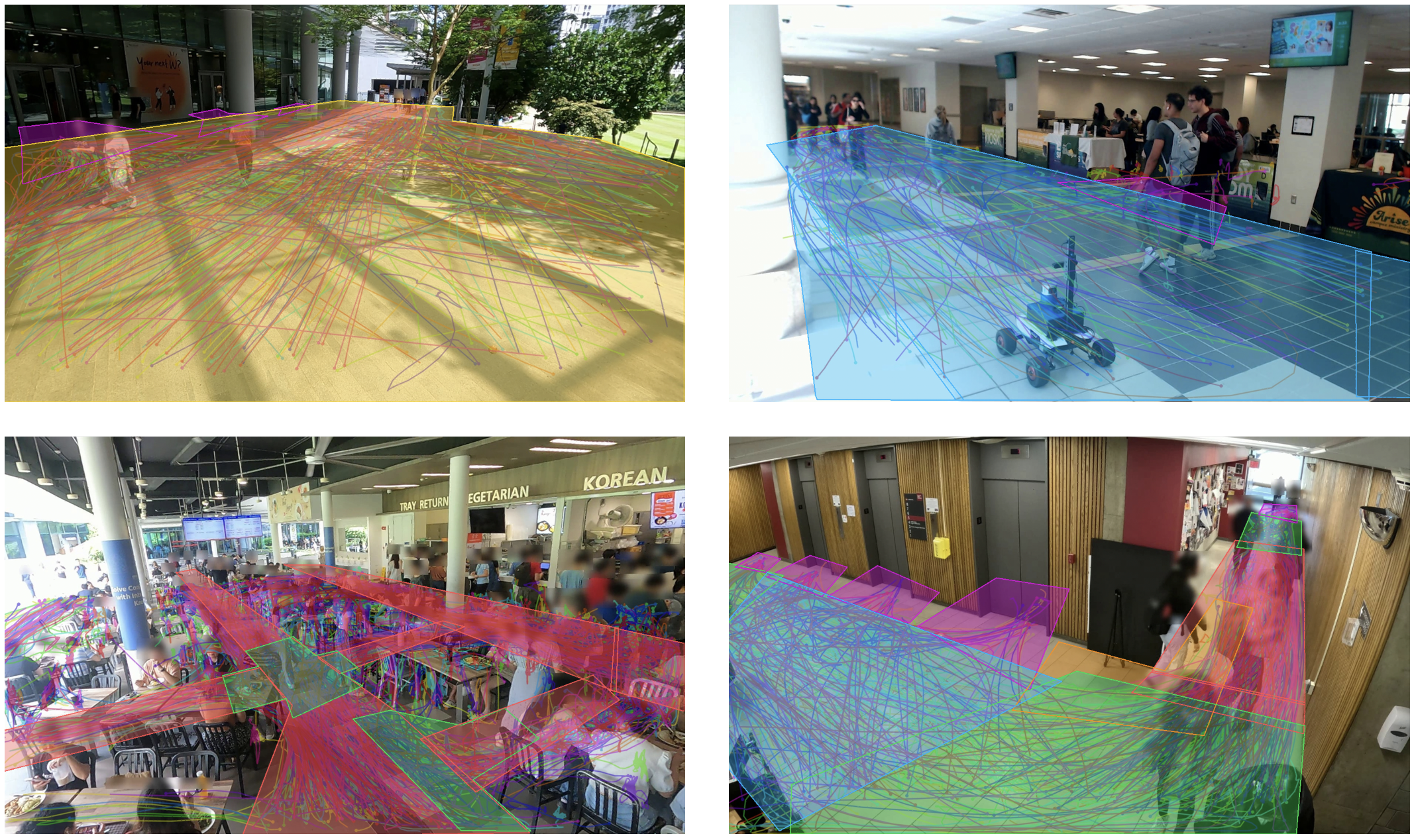}
    \caption{Example BEV tags overlaid on top of pedestrian trajectories. Each color represents a different semantic tag. The yellow areas are open spaces, the blue areas are wide corridors, the red areas are narrow corridors, the green areas are intersection areas, and the orange areas are blind corner areas. The colored lines show the annotated trajectories in the corresponding recording session.}
    \label{fig:semantic_tag}
    \vspace*{-1.0\baselineskip}
\end{figure}
BEV data offers an overhead view of the data collection site, and similar to the SiT dataset~\citep{bae2023sit}, we perform semantic tagging to characterize the various location types within a single scene. As shown in Fig.~\ref{fig:semantic_tag}, data collection areas are segmented and categorized using the same labels as the location tags in the robot data scenario tagging (i.e., `narrow corridors', `wide corridors', `intersections', `blind corners', `open spaces', and `entries/exits'). These semantic annotations can be used to analyze pedestrian trajectories within and across different scene contexts and may be used as auxiliary information when training models that incorporate human motion behavior.

\section{Using the \ACRONYM Dataset}
\subsection{Onboard Robot Data}
The \ACRONYM dataset's onboard robot data is provided in two complementary formats to support a wide range of research applications: raw ROS2 bag files for full-fidelity playback and processing, and pre-processed, uniformly sampled data in human-readable format at 4 Hz. 
The preprocessed dataset is organized hierarchically: each of the contributing teams has a dedicated folder containing all trajectories collected by that team. Within each team’s directory, data is separated into folders, each corresponding to a single recorded trajectory. These are obtained by segmenting a continuous recording session with ``Begin", ``End", and ``Discard" signals as shown in the Fig \ref{fig:data_splitting}. The folder names encode metadata such as location and recording date. Each trajectory folder contains a set of subfolders, one for each sensor modality. Data across these subfolders is time-synchronized and obtained by sampling the ROS2 bag files at a 4Hz rate. Files are named according to the index of their timestamp in the sequence of sampled timestamps (at $4$Hz) in {index}.{extension} format to ease accessing cross-modality data with time synchronization. 
For each trajectory, the following data streams are available, provided in separate files/folders:
\begin{itemize}
    \item \textbf{Odometry}: Provided in the form of CSV files containing 10 headers corresponding to the timestamp, X-Y position with respect to the odom frame, yaw, and the velocities along these dimensions.
    \item \textbf{Egocentric RGB images}: From the robot’s onboard cameras, stored as JPEG files. 
    Additionally, we also provide pedestrian bounding box detections corresponding to the RGB image frames in JSON format.
    \item \textbf{LiDAR point clouds}: Point Clouds from onboard LiDARs are provided in PCD format for easy processing and visualization.
    \item \textbf{Robot Speech}: Instances where robot speech was used are enumerated with a 3-column CSV file for each data subset containing the timestamp, trajectory name, and string corresponding to the utterance from the robot's speaker.
\end{itemize}
Along with this, we provide the full ROS2 bag files containing all recorded topics in their original message formats. Each bag file corresponds to a single trajectory and is provided in a separate folder with a similar folder structure as the processed data. 

We also provide a CSV file containing the trajectory name and the corresponding scenario tags as well as static transforms, extrinsics, and intrinsics information. 
Researchers can filter trajectories with these scenario tags to benchmark performance or train policies for specific social navigation scenarios (e.g., high-density crowds, intersections, or passing conversational groups).

\begin{figure*}
    
    \begin{subfigure}[b]{0.33\linewidth}
        \centering
        \includegraphics[width=\textwidth]{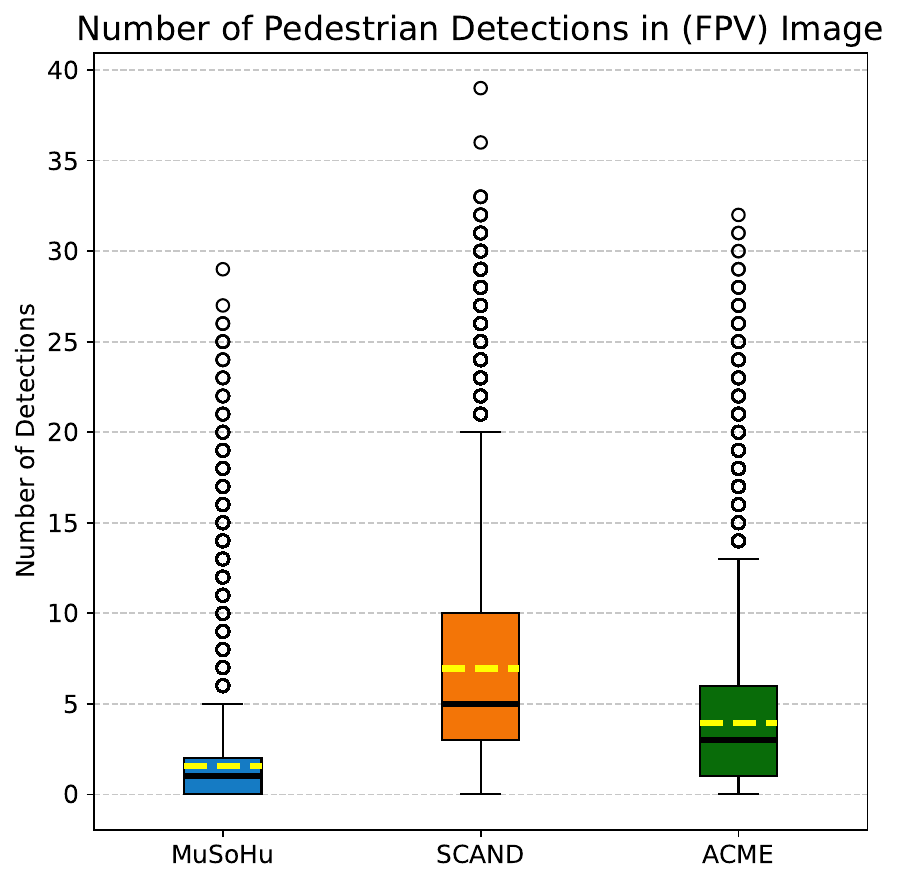}
        \caption{}
    \end{subfigure}
    \hfill 
    \begin{subfigure}[b]{0.33\linewidth}
        \centering
        \includegraphics[width=\textwidth]{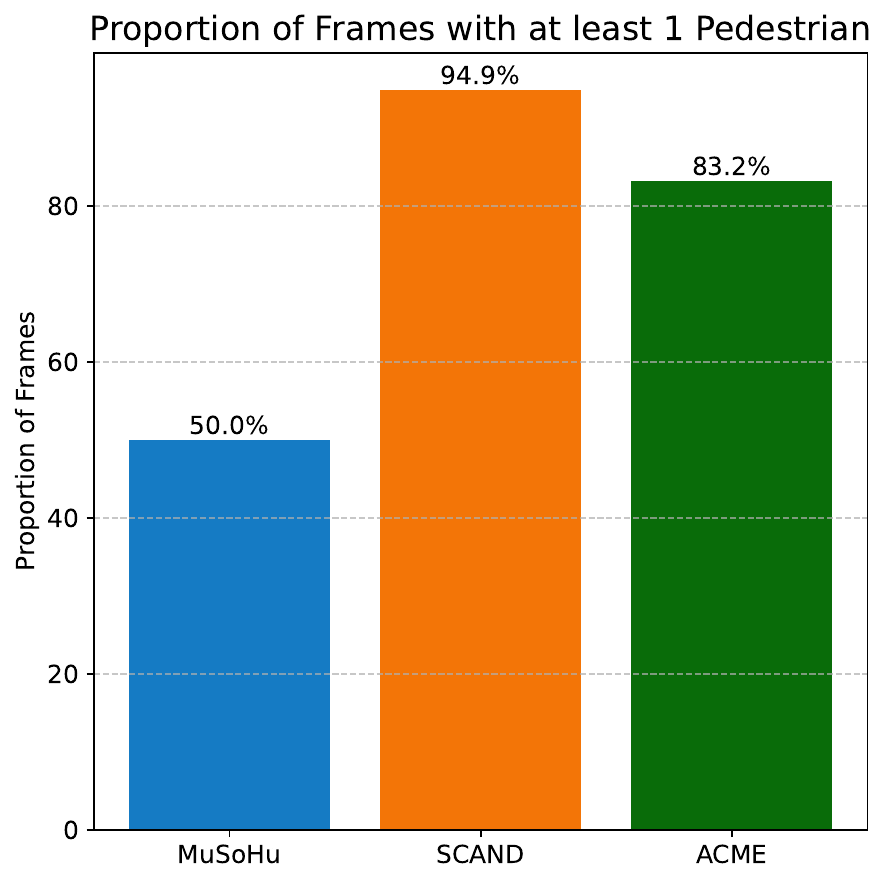}
        \caption{}
    \end{subfigure}
    \hfill 
    \begin{subfigure}[b]{0.33\linewidth}
    \centering
        \includegraphics[width=\textwidth]{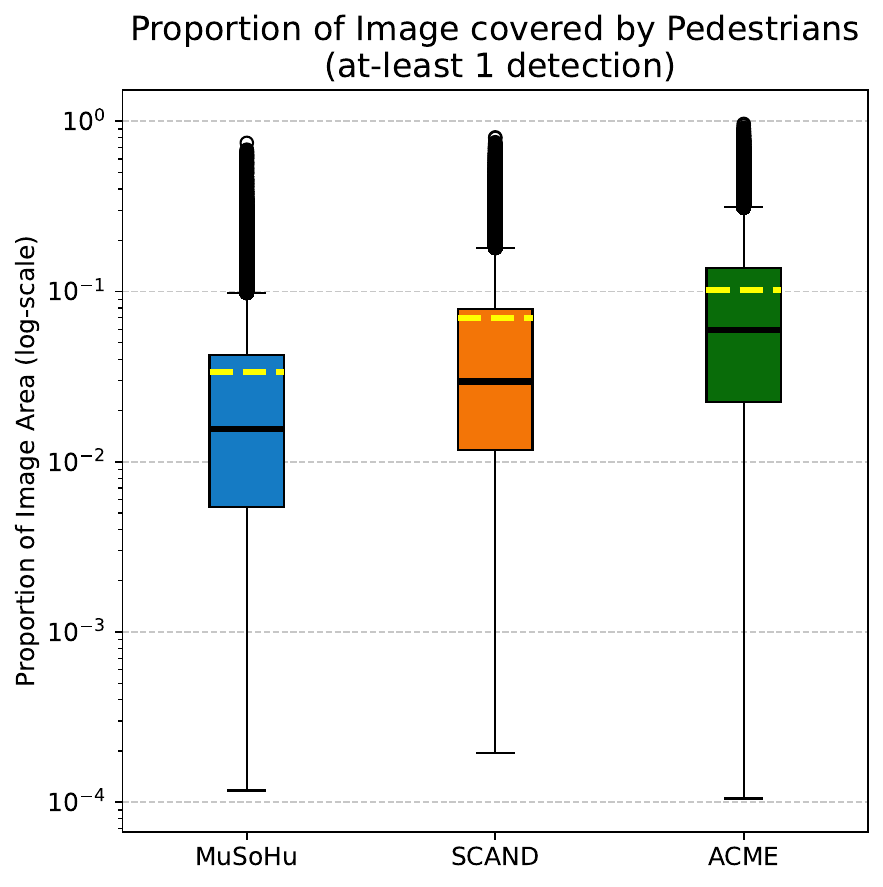}
        \caption{}
    \end{subfigure}
    \hfill 
    \begin{subfigure}[b]{0.49\linewidth}
        \centering
        \includegraphics[width=0.8\textwidth]{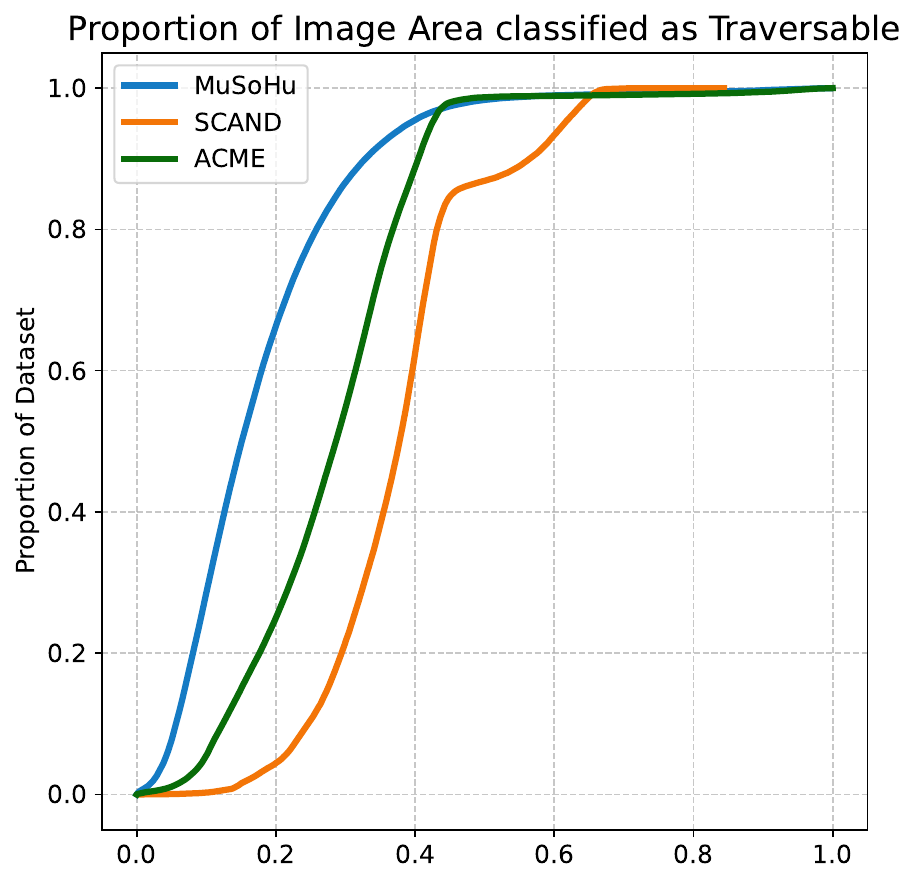}
        \caption{}
    \end{subfigure}
    \hfill 
    \begin{subfigure}[b]{0.49\linewidth}
        \centering
        \includegraphics[width=0.8\textwidth]{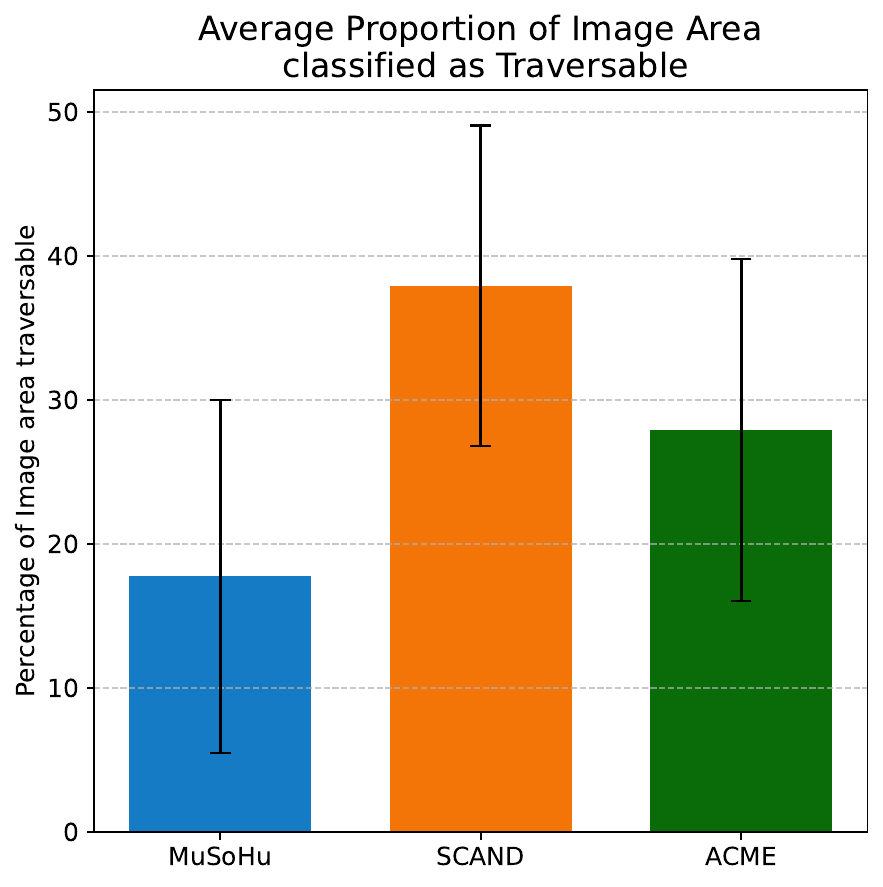}
        \caption{}
    \end{subfigure}

    \hfill
    \begin{subfigure}[b]{0.48\linewidth}
        \includegraphics[width=\textwidth]{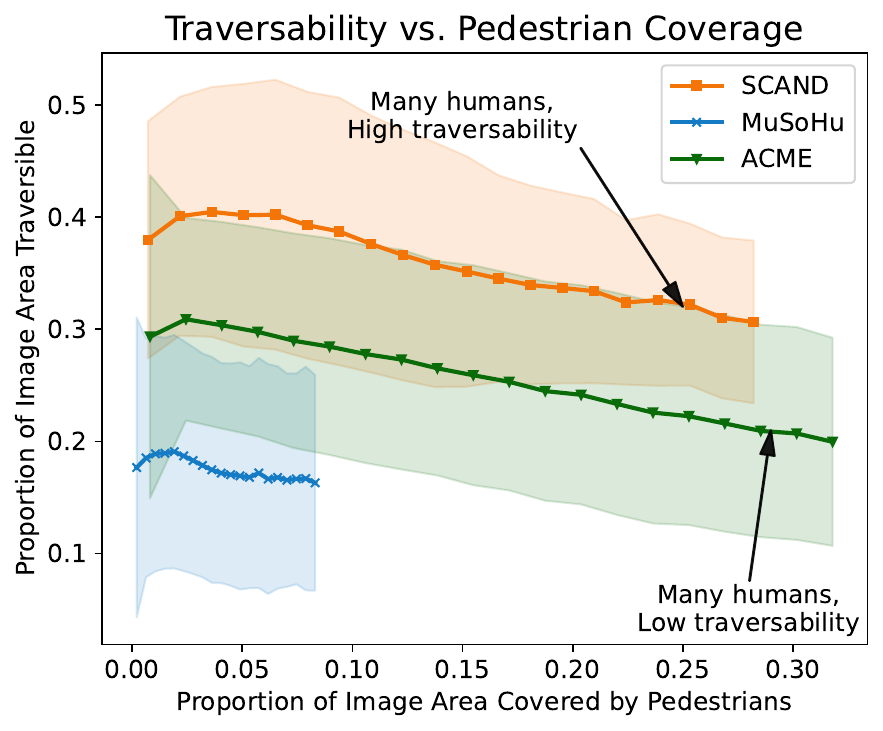}
        \caption{}
    \end{subfigure}
    \hfill 
    \begin{subfigure}[b]{0.48\linewidth}
        \includegraphics[width=\textwidth]{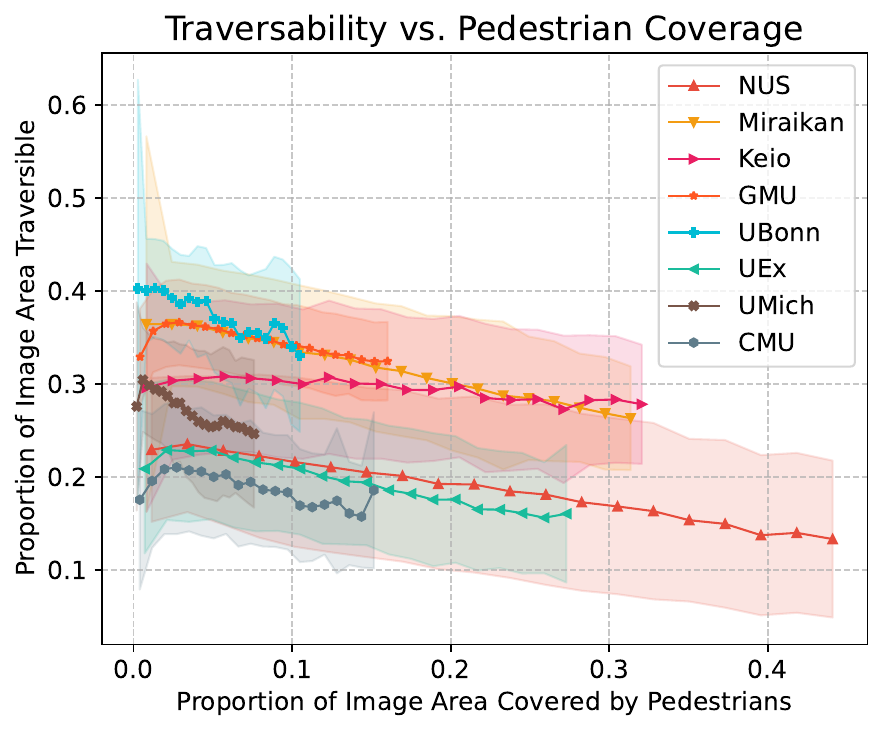}
        \caption{}
    \end{subfigure}
    \caption{Pedestrian density and traversability analysis of scenes captured from egocentric view across \ACRONYM and 2 recent Social navigation datasets: SCAND \citep{karnan2022scand} and MuSoHu \citep{nguyen2023toward}.}
    \label{fig:combined_ped_genie_analysis}
\end{figure*}

\subsection{BEV Camera Data} 
The \ACRONYM dataset's BEV data mainly consists of annotated human trajectories in 10Hz pixel coordinates and the homography matrices that transform the pixel coordinates into ground plane metric coordinates. They are saved as toml and txt files, respectively. Similar to the TBD dataset \citep{wang2024tbd}, the human trajectories were verified by the author group, and focused mainly on the moving pedestrians. Each trajectory contains an ID, a starting frame number, and a series of trajectory coordinates. The semantic scene tags are additionally provided in mask image formats as shown in Fig.~\ref{fig:semantic_tag}. These tags can be adjusted by using our open-source tagging tool. Lastly, BEV RGB videos are provided, adhering to each institution's anonymization requirements. Metadata is also included and contains information such as the BEV camera's calibrated intrinsics and distortion coefficients.

We also provide the transformed trajectories in metric coordinates on the ground plane as toml files. They are downsampled to 2.5Hz and organized into txt files following ETH~\citep{ETH} and UCY~\citep{UCY} dataset formats, so that trajectory prediction models can directly load the BEV trajectory data. To ensure the metric trajectories are clean, the transformed trajectories were further processed as follows: 
\begin{enumerate}
    \item Discard trajectories shorter than $3.2s$.
    \item In rare occasions, if there is a sudden jump in trajectories due to tracking noise (instantaneous velocity $>5m/s$), perform a linear interpolation-based fix to connect to coordinates at later timestamps. If no coordinates are found, this segment is discarded.
    \item Smooth the trajectories using a smoothing window size of $5$.
    \item Label the trajectory segments that are stationary (average speed over a window of $8$ time steps is $<0.5m/s$).
    \item Limit the trajectories of the pedestrians to be within $(\pm25m, \pm25m)$ of the poster-sized tag. This mainly affected UBonn and NUS because some of the data collection areas were large, and the cameras were pointed at a narrower angle, so it was difficult to obtain accurate positions for pedestrians who were far away. 
\end{enumerate}
The BEV trajectory data analyzed in the evaluation sections were processed using these procedures.

\externaldocument{utility}
\section{Dataset Analysis: Social navigation scenes from onboard-robot data}
In this section,  we analyze on-board robot data collected in \ACRONYM and compare it to prior datasets focused on social navigation, specifically SCAND \citep{karnan2022socially} and MuSoHu \citep{nguyen2023toward}. We envision \ACRONYM to be useful for learning social navigation robot policies as well as pedestrian trajectory prediction models, and therefore aim to capture scenarios that prove to be challenging for current off-the-shelf navigation models. Our metrics for comparison focus on the complexity and diversity of data captured in each dataset. ``Scenario complexity" is hard to quantify and highly subjective. Although there have been recent efforts to characterize scenario complexity \citep{stratton2024characterizing}, analyzing large datasets with the relevant factors that constitute such metrics in an automated manner remains challenging. Thus, we design metrics that can be scalably computed on any dataset. Our characterization of scenario complexity comprises 3 factors that address the inherent multidimensional nature of social navigation scenarios: 
\begin{enumerate}
    \item Pedestrian Density: How crowded is the scene that the robot finds itself in?
    \item Traversability: How constrained is the robot's motion due to pedestrians/other spatial constraints?
    \item Degree of Social Compliance (\cite{raj2024rethinking}): How would an off-the-shelf geometric planner behave in the situation the robot finds itself in? How different is this trajectory from the expert ``socially compliant" trajectory?
\end{enumerate}
\begin{figure}
    \centering
    \begin{subfigure}[b]{0.48\linewidth}
        \includegraphics[width=\textwidth]{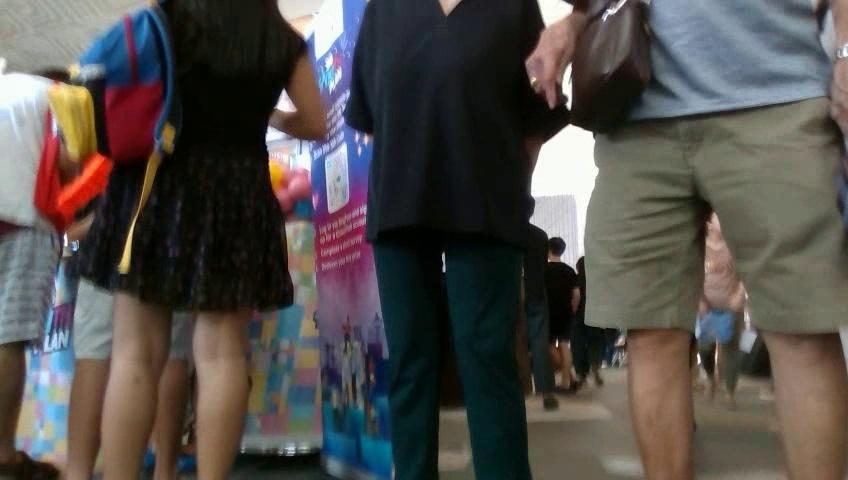}
    \end{subfigure}
    \hfill 
    \begin{subfigure}[b]{0.48\linewidth}
        \includegraphics[width=\textwidth]{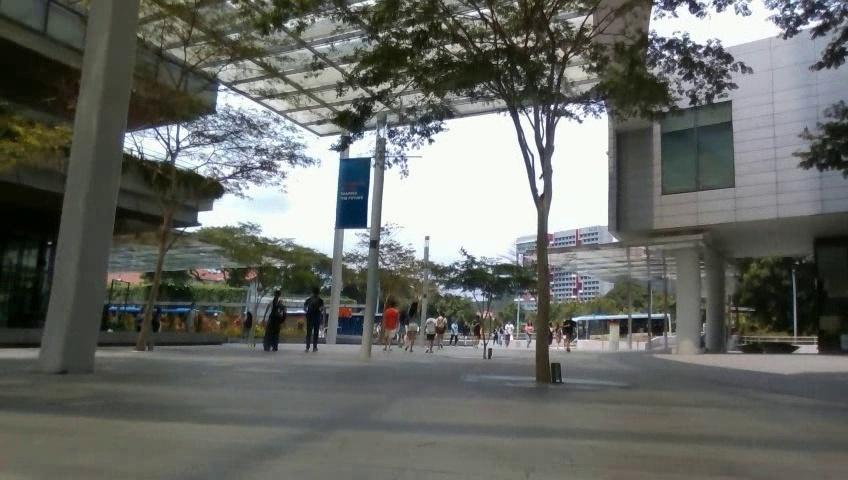}
    \end{subfigure}
    \caption{Navigation scenarios from \ACRONYM  with the same number of detected pedestrians but presenting vast differences in the navigation challenge faced.}
    \label{fig:ped_density_difference}
    \vspace*{-1.0\baselineskip}
\end{figure}
\subsection{Pedestrian Density}
We use the same off-the-shelf Multi-Object Tracking model used for anonymizing our data to generate pedestrian detections on egocentric images for SCAND \citep{karnan2022socially} and MuSoHu \citep{nguyen2023toward}. We characterize pedestrian density with two metrics: 1) Number of pedestrians detected in the image, and 2) Proportion of the image covered by pedestrians. The latter is an important statistic to identify how close people are to the robot (as a proxy when depth information and intrinsic camera parameters are unavailable), which is indicative of the crowd density the robot is navigating through. For example, in Fig \ref{fig:ped_density_difference}, although both the images have 8 (detected) pedestrians in total, the scenario on the left offers a much greater challenge to navigate safely as opposed to the scenario on the right owing to a larger number of pedestrians close to the robot. Fig \ref{fig:combined_ped_genie_analysis}(b) shows that \ACRONYM captures far fewer scenes with no humans and remains competitive with SCAND (roughly 1/4th the size of our dataset by duration) in terms of the number of pedestrians captured per scene. Notably, while SCAND captures a larger number of scenes with a high pedestrian count (Fig. \ref{fig:combined_ped_genie_analysis}(a)), \ACRONYM captures a larger proportion of scenes with pedestrians closer to the robot (Fig.\ref{fig:combined_ped_genie_analysis}(c)). 

\subsection{Traversability} The area available for the robot to safely traverse the environment in the presence of humans can indicate the complexity of scenarios captured. Larger traversable regions generally provide more feasible paths around pedestrians, while constrained regions increase the likelihood of close-proximity interactions and navigation decisions that require social awareness.
To quantify this aspect, we process the SCAND, MuSoHu, and \ACRONYM datasets using a finetuned SAM2 model (\cite{wang2024genie}) to predict a \textit{traversability mask} on the egocentric RGB images.
Fig.~\ref{fig:combined_ped_genie_analysis}(c,d) compares the proportion of image area classified as traversable across \ACRONYM, SCAND, and MuSoHu. SCAND exhibits the largest traversable regions on average, suggesting that many of its scenes provide relatively open planning spaces. In contrast, MuSoHu is shifted toward lower traversability values, indicating more spatially constrained scenes. \ACRONYM lies between these two datasets: its scenes contain less free navigation space than SCAND, but more than MuSoHu. This suggests that \ACRONYM captures a broad range of moderately constrained navigation settings, where the robot often has feasible alternatives but still encounters reduced planning space that can require socially aware decision-making. 
Furthermore, we analyze the relationship between pedestrian coverage (the pixel-wise proportion of egocentric images occupied by pedestrians) and traversability. While raw pedestrian counts can be misleading in open spaces, pedestrian coverage serves as a more reliable proxy for agent proximity and the resulting obstruction of the robot’s path.

As illustrated in Fig.\ref{fig:combined_ped_genie_analysis} (f), we observe a clear inverse correlation: as pedestrians occupy more of the visual field (indicating closer proximity or higher local density), the available traversable area decreases. Notably,~\ACRONYM captures a significantly higher density of `socially constrained' scenarios compared to SCAND and MuSoHu (lower-right regions in Fig \ref{fig:combined_ped_genie_analysis} (f)). While MuSoHu contains many low-traversability scenes, these are primarily narrow, empty environments with few pedestrians. In contrast, \ACRONYM focuses on the long-tail of social navigation—scenes where the robot must negotiate constrained spaces, specifically while navigating through or around human crowds

\subsection{Robot Speech}
\begin{figure}
    \centering
    \begin{subfigure}{0.5\textwidth}
        \includegraphics[width=\linewidth]{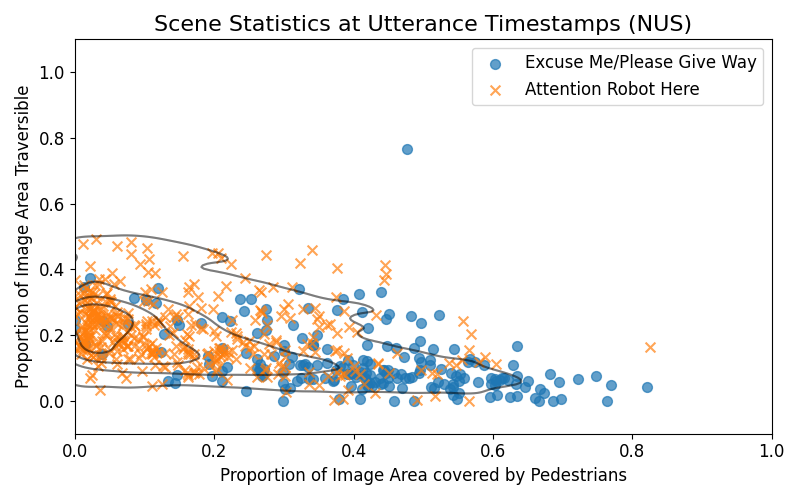}
    \caption{Scenes with Robot speech utterances in NUS data}
    \label{fig:nus_trav_ped_speech}
    
    \end{subfigure}
    
    \hfill
    \begin{subfigure}{0.5\textwidth}
        \includegraphics[width=\linewidth]{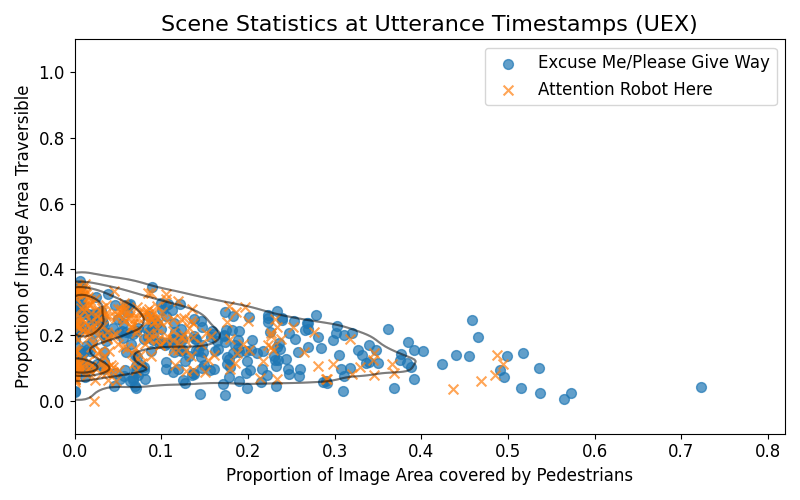}
    \caption{Scenes with Robot speech utterances in UEx data}
    \label{fig:uex_trav_ped_speech}
    \end{subfigure}
    \hfill
    \begin{subfigure}{0.5\textwidth}
        \includegraphics[width=\linewidth]{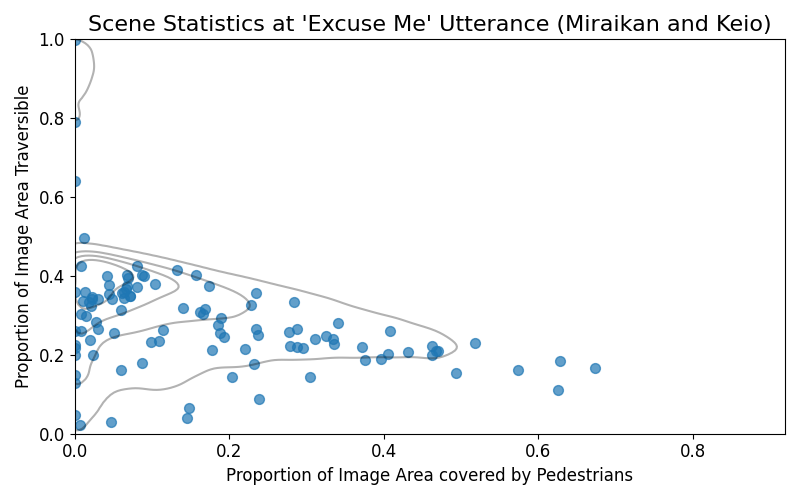}
    \caption{Scenes with speech utterances in Miraikan \& Keio data}
    \label{fig:japan_trav_ped_speech}
     \end{subfigure}
    \label{fig:speech_trav_ped}
    \vspace*{-1.0\baselineskip}
    \caption{Robot speech usage in the NUS, UEx, Miraikan, and Keio subsets. Each marker corresponds to a frame in which a robot speech utterance was issued, plotted by pedestrian density and traversability at that timestamp. Gray contour lines show the overall distribution of pedestrian density and traversability within the corresponding data subset.}
    \label{fig:speech_scatter}
    \vspace*{-1.0\baselineskip}
\end{figure}

Trajectories collected in NUS, UEx, Miraikan, and Keio also contain annotations of robot speech usage, indicating when and where robot speech was used during crowd navigation. We discovered that the robot operation method impacts robot speech behavior during data collection and thus makes a distinction between the \textit{teleoperated} NUS and UEx robots and the \textit{human-operated} Miraikan and Keio suitcase robots.

\subsubsection{Tele-operated robots}
Recall that the NUS and UEx robots were configured with onboard speakers and could actively play one of 3 utterances: `Excuse Me', `Attention! Robot Here' and `Please Give Way'. In the NUS and UEx subsets, we found `Excuse Me' and `Please Give Way' were used under similar interaction conditions, and thus we combined them into a single category. Post-collection feedback from teleoperators suggests that the utterances correspond to two broad interaction functions:
\begin{itemize}
    \item \textbf{``Excuse Me/Please Give Way"}: When the robot's path was hindered by pedestrians or the robot moved in close proximity to/overtook a person or group of people. This typically occurred in crowded scenes, where the robot needed to request passage to avoid collision. 
    \item \textbf{``Attention Robot Here"},: When pedestrians needed to be alerted of the robot's presence, especially in situations of low visibility (e.g., doorways/blind corners) or the robot overtakes/passes by a distracted pedestrian (e.g., looking at a phone) and there was a possibility of collision. In these cases, speech served primarily as an awareness cue rather than an explicit request for passage.
\end{itemize}
These usage patterns are also reflected in the FPV scenes corresponding to utterance timestamps. Fig \ref{fig:speech_scatter} compares the proportion of image area inferred as traversable against the proportion of image area occupied by pedestrians at speech timestamps. In both NUS and UEx, `Attention Robot Here' is used at relatively lower pedestrian densities, whereas `Excuse Me/Please give way' is used in more crowded scenes for space negotiation. 
The plots also reveal an embodiment-dependent difference between the NUS and UEx robots. In the NUS subset, `Attention Robot Here' is distributed over a wider range of pedestrian densities and traversability values. This is consistent with the NUS platform being a small quadruped at a maximum height of $40$~cm from the ground, making it less visually salient in crowds than the $1.33$~m tall UEx shadow robot. 

\subsubsection{Human Operated Robots}
For the suitcase-robot trajectories in Miraikan and Keio, the human operators pressed a button on the suitcase handle interface whenever they verbally said `Excuse Me' during a trajectory recording. Because the human operators were physically connected to the robots and were adept at crowd navigation, pedestrians respond primarily to the accompanying human rather than the inconspicuous suitcase as an autonomous social actor. As a result, `Excuse Me' occurred only in the context of causing disturbance to other pedestrians from the judgment of the human operator. In contrast, when the robots were teleoperated in NUS and UEx, the teleoperators were less in control and executed speech more often. As a result, even though the combined duration of the Miraikan and Keio data is comparable to that of the combined NUS and UEx subsets, speech usage is substantially lower in the suitcase-robot data as shown in Fig \ref{fig:japan_trav_ped_speech}.

In Miraikan and Keio, when the accompanying human operators caused disturbance to other pedestrians and uttered `Excuse Me', the robots often were not in the immediate vicinity of any other pedestrian. This is because disturbances in Miraikan and Keio often took the form of causing other pedestrians to change course before getting close to them, or cutting through empty space where pedestrians were interacting with each other or with objects in the environment (the reasonable way to drive the robot in high-density situations based on the operators' judgment). During these disturbance events, the robot may not be physically close to other pedestrians. This may be the underlying reason that the data in Fig \ref{fig:japan_trav_ped_speech} on Miraikan and Keio concentrates towards higher traversable area proportions and lower pedestrian image area proportions.

Finally, we emphasize that the quantities in Fig \ref{fig:speech_scatter} are computed in FPV image space with trained models for detecting pedestrians and traversability. They should not be interpreted as direct metric measurements of crowd density or navigable free space. These quantities are affected by camera intrinsics, extrinsics, and segmentation quality. With respect to speech usage patterns, the most reliable interpretation is not that a universal visual threshold determines when speech is used, but rather that robot-speech is a situated, embodiment-dependent navigation action whose use correlates with visually constrained and socially interactive scenes.
\subsection{Comparison to a Geometric Planner}
\begin{figure*}
     \centering
     \begin{subfigure}[b]{0.33\textwidth}
         \centering
         \includegraphics[width=\textwidth]{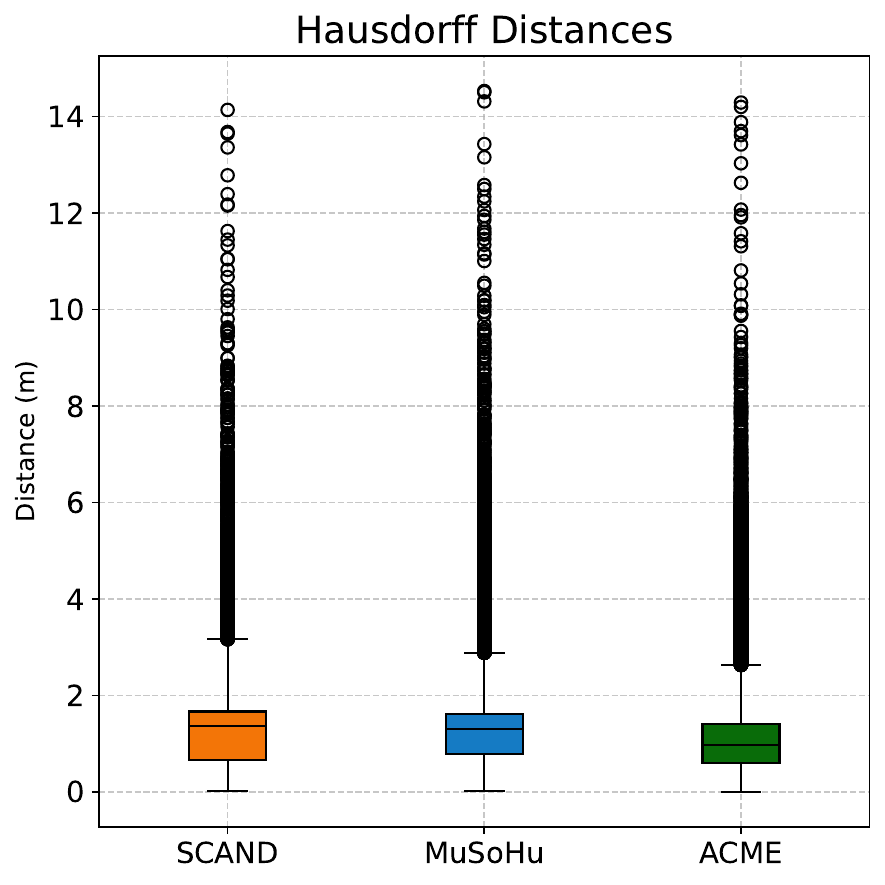}
     \end{subfigure}
     \hfill
     \begin{subfigure}[b]{0.33\textwidth}
         \centering
         \includegraphics[width=\textwidth]{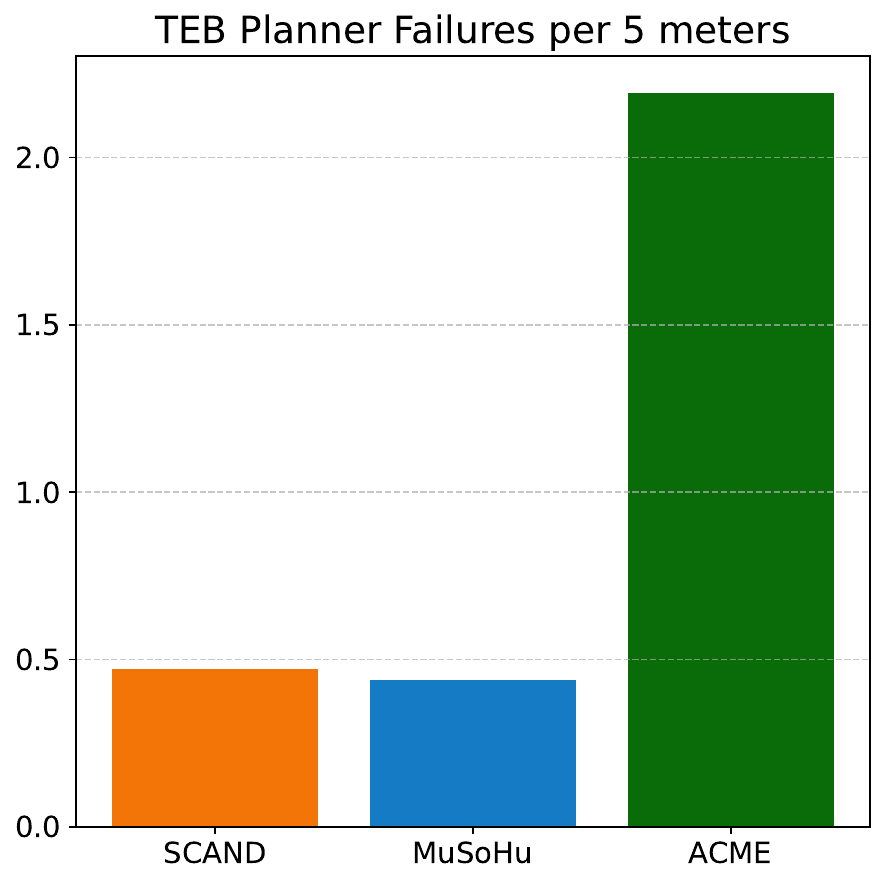}
     \end{subfigure}
     \hfill
     \begin{subfigure}[b]{0.33\textwidth}
         \centering
         \includegraphics[width=\textwidth]{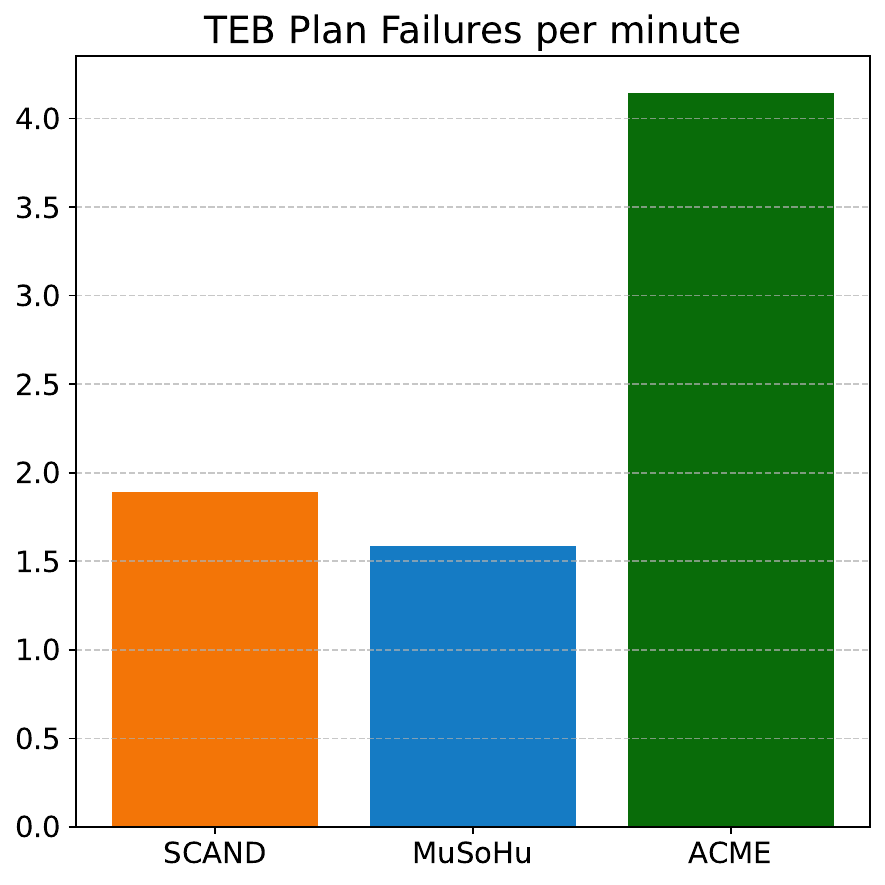}
     \end{subfigure}
     \caption{Although the distribution of Hausdorff distances between the expert trajectory and TEB remains similar across datasets, the higher planner failure rates showcase that off-the-shelf planners have more difficulty navigating in the scenes captured in \ACRONYM compared to other datasets.}
     \label{fig:move_base_results}
\end{figure*}

\cite{raj2024rethinking} defined ``social-compliance" of motion planners using the Hausdorff distance between the global plan generated by the planner and the expert trajectory. We hypothesize that our dataset, with its focus on capturing challenging scenarios, would reveal more situations where such context-and-norm-unaware planners would be deemed socially noncompliant. We perform an analysis identical to that of \cite{raj2024rethinking}, by sampling future odometry goals (5m ahead of the robot's current position at a 1Hz rate) and measure the Hausdorff distance between the plan generated by move\_base (in contrast to \cite{raj2024rethinking}, we use the TEB local planner \citep{Rosmann2017-in} to emulate real-world testing conditions) and the expert trajectory. Our experiments revealed an interesting phenomenon: as shown in Fig \ref{fig:move_base_results}.a, the undirected Hausdorff distance between the TEB planner and the expert trajectory remains similar across all datasets. However, analysis of the failure rate of the planners reveals a clear trend (Fig. \ref{fig:move_base_results} b): scenes in the \ACRONYM dataset are far more challenging to navigate than SCAND and MuSoHu, resulting in 2x more failures per minute and 4x per 5 meters of data compared to the dataset with the second highest failure rate (SCAND). We verify that the primary mode of planner failure across the dataset is the presence of pedestrians at the sampled goal position for the planner, thus directly correlating with the crowd density that the robot must navigate through.

\subsection{Benchmarking Vision Navigation models}
We showcase the utility of \ACRONYM for training and testing general-purpose navigation agents by benchmarking two SoTA navigation models. Specifically, we focus on models designed for real-world vision navigation with zero-shot or few-shot transfer to novel environments and embodiments, namely ViNT \citep{shah2023vint} and NoMAD \citep{sridhar2024nomad}. We evaluate model performance based on alignment to the expert trajectories with Final Destination Error (FDE), Average MSE on waypoints, and MAOE as proposed in \cite{liu2025citywalker}, as well as Total AOE (aggregate orientation error across predicted waypoints). For reference values, we also list performance on the SCAND and MuSoHu datasets. Table \ref{tab:nav_metrics} shows the results of the benchmark. As expected, both models perform well on the SCAND (since ViNT and NoMAD are trained on SCAND data as well) and relatively poorly on the MuSoHu and \ACRONYM dataset. While NoMAD's performance on the MuSoHu and \ACRONYM dataset is very similar, ViNT performs better on MuSoHu than on \ACRONYM. The ease of using our dataset for such benchmarks, paired with the location and embodiment diversity, could bolster future work to investigate scenarios of failure and improve foundational navigation models to traverse human-inhabited spaces.
\begin{figure*}
     \centering
     \begin{subfigure}[b]{0.33\textwidth}
         \centering
         \includegraphics[width=\textwidth]{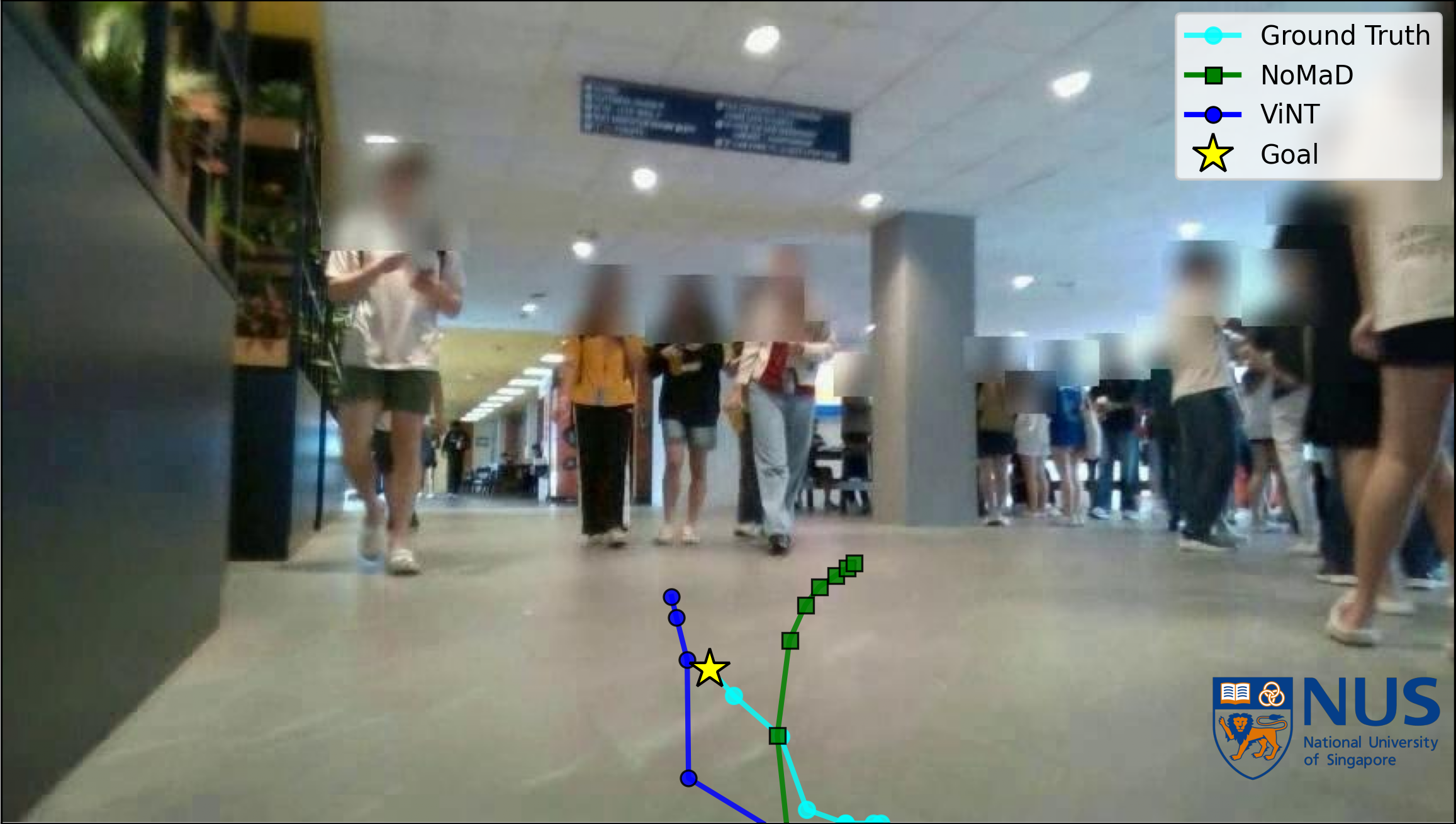}
     \end{subfigure}
     \hfill
     \begin{subfigure}[b]{0.33\textwidth}
         \centering
         \includegraphics[width=\textwidth]{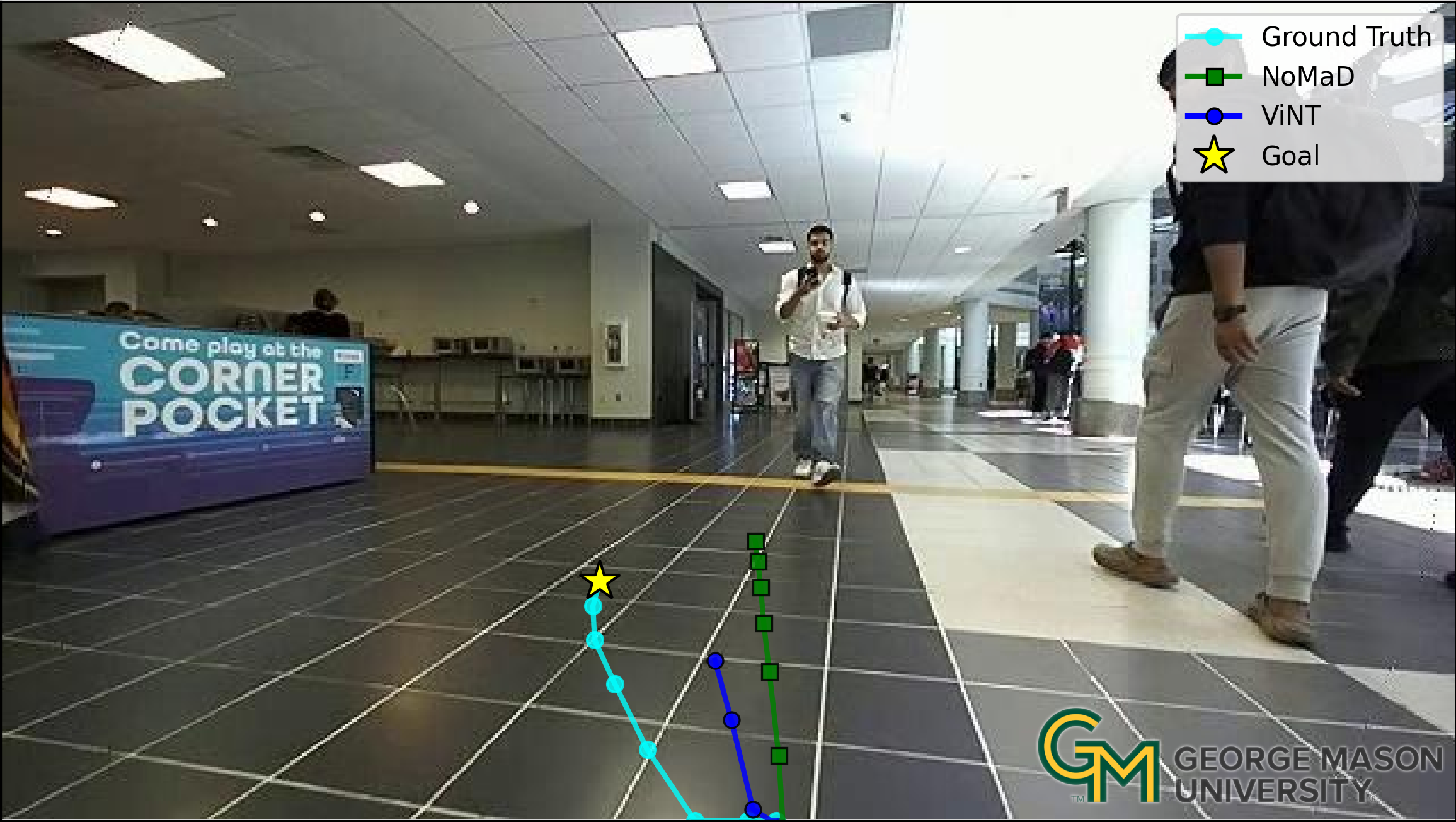}
     \end{subfigure}
     \hfill
     \begin{subfigure}[b]{0.33\textwidth}
         \centering
         \includegraphics[width=\textwidth]{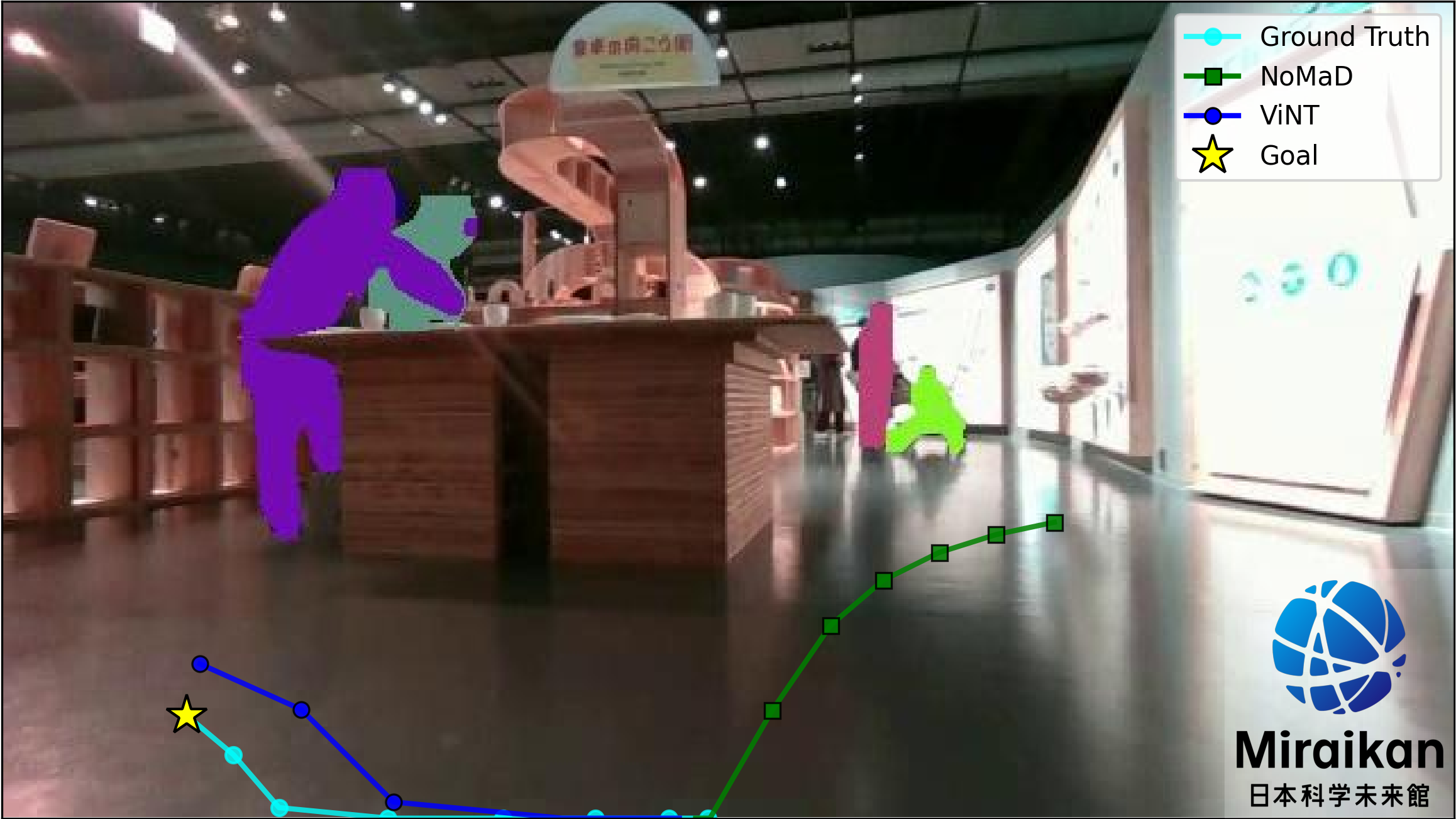}
     \end{subfigure}
     \caption{Qualitative Results: Trajectories generated by SoTA foundational navigation models compared with the ground truth trajectories.}
     \label{fig:vint_nomad_results}
\end{figure*}

\begin{table*}[!t]
\centering
\caption{Performance Benchmark of SoTA foundational navigation  models }
\label{tab:nav_metrics}
\vspace{0.5em}

\footnotesize
\renewcommand{\arraystretch}{1.2}

\begin{tabular*}{\textwidth}{@{\extracolsep{\fill}} ll
S[table-format=1.3(2)]
S[table-format=1.3(2)]
S[table-format=1.3(3)]
S[table-format=2.3(5)]
S[table-format=2.3(5)]
S[table-format=3.3(6)] @{}}
\toprule
& & {\textbf{FDE (m)}} & {\textbf{MSE}} & {\textbf{Cosine Similarity}} & {\textbf{MAOE ($\degree$)}} & {\textbf{Mean AOE ($\degree$)}} & {\textbf{Total AOE ($\degree$)}} \\
\midrule
\multirow{3}{*}{\textbf{ViNT}}
& \ACRONYM & 0.85 \pm 0.60 & 0.24 \pm 0.30 & 0.95 \pm 0.23 & 12.58 \pm 26.51 & 8.24 \pm 19.93 & 41.22 \pm 99.64 \\
 & SCAND & 0.46 \pm 0.52 & 0.11 \pm 0.23 & 0.98 \pm 0.13 & 5.48 \pm 15.47 & 3.68 \pm 11.31 & 18.41 \pm 56.53 \\
 & MuSoHu & 0.83 \pm 0.49 & 0.21 \pm 0.34 & 0.96 \pm 0.18 & 14.39 \pm 21.25 & 10.16 \pm 16.26 & 50.80 \pm 81.30 \\
 \midrule
 \multirow{3}{*}{\textbf{NoMaD}} & \ACRONYM & 1.44 \pm 1.00 & 0.63 \pm 0.77 & 0.93 \pm 0.22 & 20.68 \pm 31.35 & 11.74 \pm 19.75 & 93.88 \pm 158.00 \\
 & SCAND & 0.79 \pm 0.92 & 0.30 \pm 0.59 & 0.98 \pm 0.12 & 6.86 \pm 16.39 & 4.11 \pm 10.73 & 32.89 \pm 85.83 \\
 & MuSoHu & 1.53 \pm 0.88 & 0.63 \pm 0.77 & 0.94 \pm 0.19 & 18.83 \pm 26.10 & 11.80 \pm 17.27 & 94.42 \pm 138.16 \\
\bottomrule
\end{tabular*}
\end{table*}

\section{Dataset Analysis: Pedestrian Trajectory Analysis from BEV data}
\begin{table}[!t]
\renewcommand{\arraystretch}{1.2}
\caption{Dataset statistics comparison, for datasets with human-verified trajectories grounded in metric space.}
\label{tab:bev_stats}
\centering
\resizebox{\columnwidth}{!}{
\begin{tabular}{>{\centering\arraybackslash}m{2.6cm}|>{\centering\arraybackslash}m{1.8cm}>{\centering\arraybackslash}m{2.2cm}>{\centering\arraybackslash}m{1.5cm}}
\toprule
\textbf{Datasets} & \textbf{Duration} & \textbf{\# Trajectories} & \textbf{Freq (Hz)}\\
\midrule

\shortstack{ETH\\{\scriptsize \cite{ETH}}}
& 25 min & 650 & 15 \\

\shortstack{UCY\\{\scriptsize \cite{UCY}}}
& 16.5 min & 786 & 2.5 \\

\shortstack{Town Centre\\{\scriptsize \cite{towncentre}}}
& 5 min & 157 & 2.5 \\

\shortstack{WildTrack\\{\scriptsize \cite{chavdarova2018wildtrack}}}
& 200 sec & 313 & 2 \\

\shortstack{JRDB\\{\scriptsize \cite{jrdb}}}
& 62 min & $\sim$3.5K & 7.5 \\

\shortstack{TH\"OR\\{\scriptsize \cite{thor}}}
& 60+ min & 600+ & 100 \\

\shortstack{TBD\\{\scriptsize \cite{wang2024tbd}}}
& 626 min & 10.3K & 10 \\

\shortstack{SiT\\{\scriptsize \cite{bae2023sit}}}
& 9 min & 1861 & 10 \\

\shortstack{TH\"OR-MAGNI\\{\scriptsize \cite{schreiter2025thormagni}}}
& 1416 min & $\sim$10K & 100 \\

\shortstack{Bi$^3$\\{\scriptsize \cite{stratton2026bi3dataset}}}
& 630 min & $\sim$11K & 120 \\

\shortstack{\ACRONYM\\{\scriptsize (Ours)}}
& 2613 min & 72.1K & 10 \\

\bottomrule
\end{tabular}
}

\end{table}
For comparisons among trajectory prediction datasets, we first analyze basic statistics, i.e., duration, speed, and density. We then benchmark state-of-the-art trajectory prediction models to compare with the combined ETH \citep{ETH} and UCY \citep{UCY} dataset. Lastly, to examine whether our dataset captures diverse human behavior, we compare \ACRONYM's sub-datasets from different countries and semantic environment tags, using context-dependent metrics. We revealed several differences in human navigation behavior from different cultures.
Table~\ref{tab:bev_stats} shows the quantity of our \ACRONYM dataset compared to prior datasets that contain human-verified pedestrian trajectory annotations in the metric space. Compared to the second largest dataset, the TBD dataset \citep{wang2024tbd}, \ACRONYM is almost 4 times larger by duration, and contains 7 times more pedestrian trajectories. Additionally, the TBD dataset only contains data at one location.

\begin{figure}[t]
    \centering
    \captionsetup[subfigure]{justification=centering}
    \begin{subfigure}{0.485\textwidth}
        \centering
        \includegraphics[width=\textwidth]{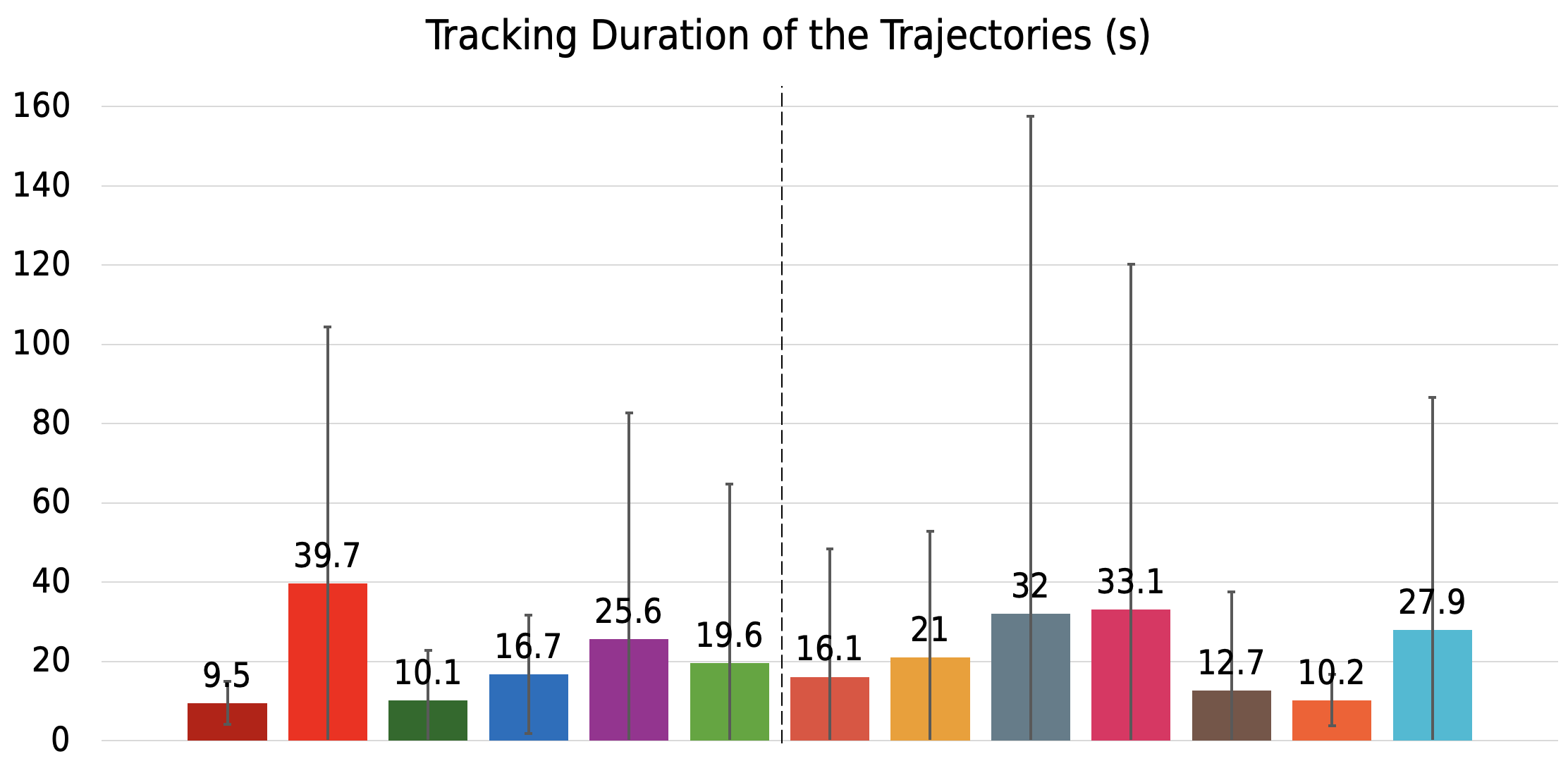}
        \caption{ }
    \end{subfigure}
    \\
    \begin{subfigure}{0.485\textwidth}
        \centering
        \includegraphics[width=\textwidth]{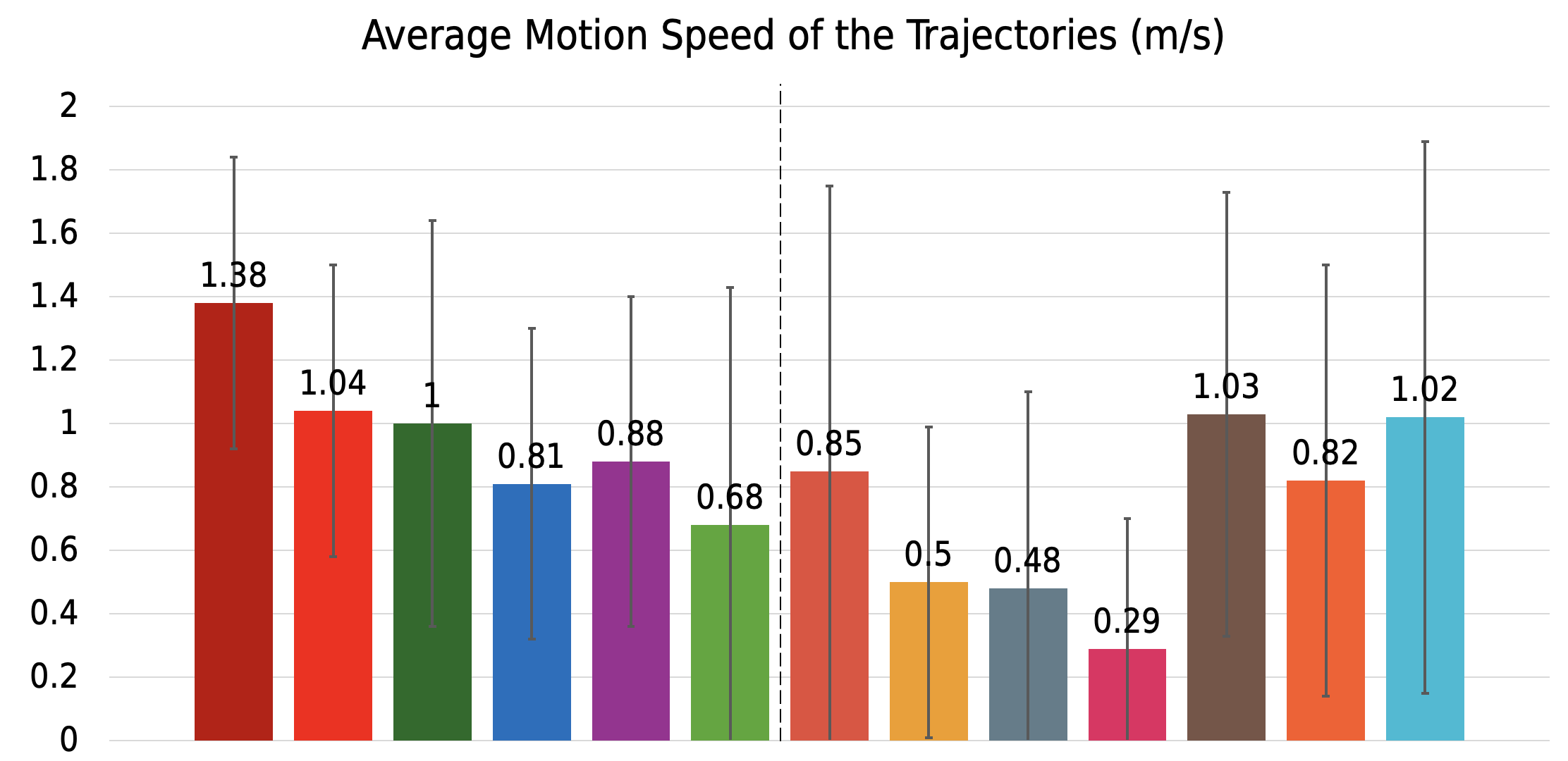}
        \caption{ }
    \end{subfigure}
    \\
    \begin{subfigure}{0.485\textwidth}
        \centering
        \includegraphics[width=\textwidth]{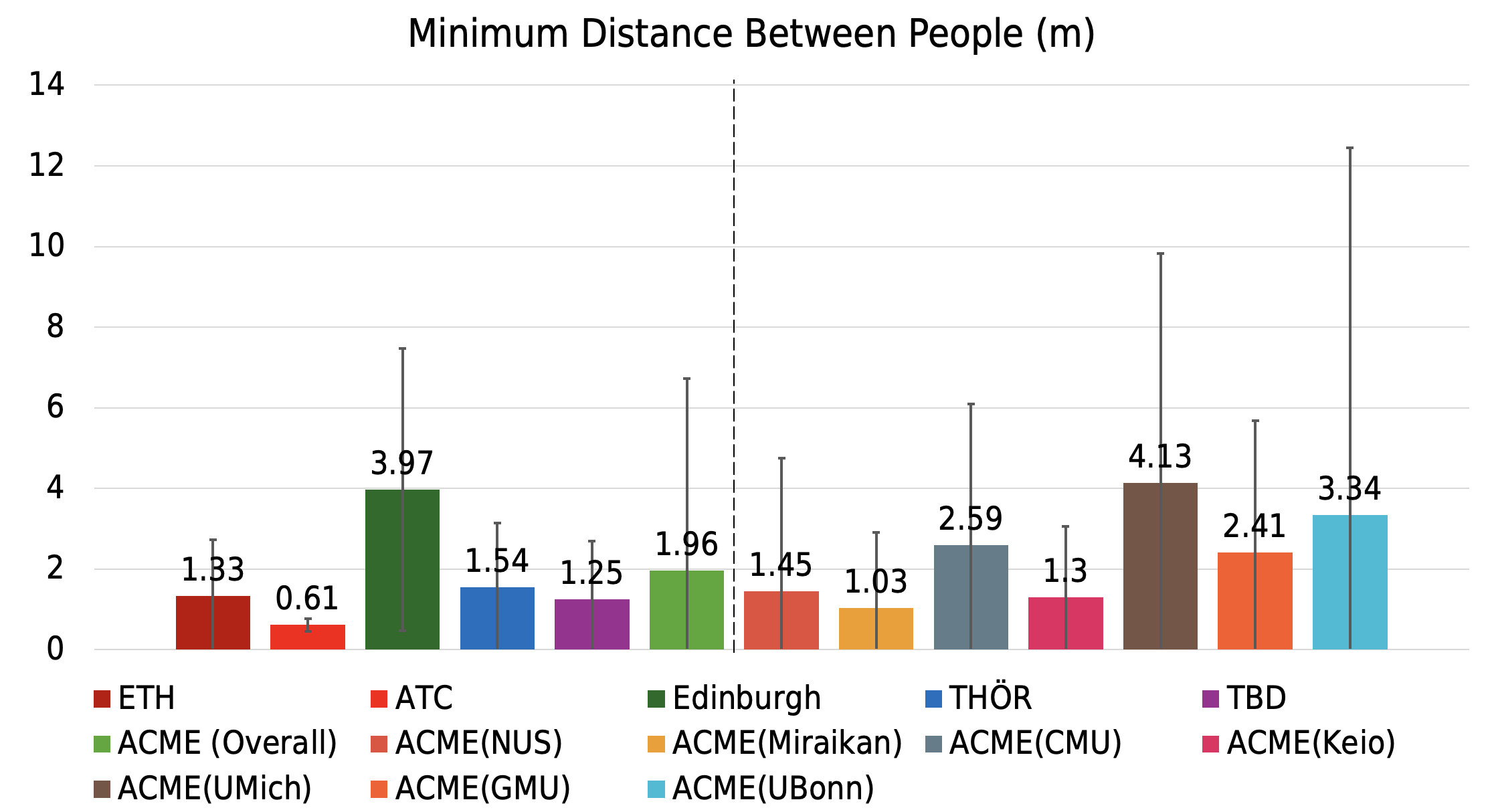}
        \caption{ }
    \end{subfigure}
    \caption{Trajectory statistics of our data compared to prior datasets with $\pm$ standard deviation. The statistics on the right side of the dashed lines are data for each of our sub-datasets. (a) duration in seconds. (b) average motion speed in meters per second. (c) minimum distance between any two pedestrians over time in meters.}
    \label{fig:bev_traj_stat}
    \vspace*{-1.0\baselineskip}
    
\end{figure}

We additionally compare the statistics of our dataset and representative datasets, extending the evaluation by \cite{thor}. In particular, we use the following metrics: (1) \textbf{Tracking Duration} ($s$): average time duration of the trajectories. (2) \textbf{Motion Speed} ($m/s$): average speed of the trajectories. (3) \textbf{Minimum Distance Between People} ($m$): minimum Euclidean distance between any two people, averaged over frames, which partially reflects overall dataset density. The perception noise metric from \cite{thor} is not included in this analysis, as it is revealed in practice that this metric is heavily influenced by the implementation of our post-processing pipeline. For example, a longer trajectory smoothing window will lower perception noise significantly, but at the cost of distorting the trajectories. And for reasons similar to \cite{wang2024tbd}, trajectory curvature is not measured.

\subsubsection{Tracking Duration} As shown in Figure~\ref{fig:bev_traj_stat}a, our dataset's mean tracking duration is not the highest, but with a large standard deviation ($\pm45.1$). The large variation is typically caused by wandering pedestrians and static pedestrians stopping for various activities, such as ordering food or having a conversation with others. These pedestrian trajectories' durations are often long and varied. Other datasets with a large presence of wandering and static pedestrians also have large variances, such as the ATC ($\pm 64.7$) and TBD ($\pm 57.1$) datasets. Among the sub-datasets, NUS, Miraikan, UMich, and UBonn contain a moderate mix of high-duration trajectories, while CMU and Keio contain a high number of high-duration trajectories.

\subsubsection{Average Speed} As shown in Figure~\ref{fig:bev_traj_stat}b, overall, the average speed in our dataset is the lowest, caused by large numbers of static pedestrians. However, our dataset has the highest standard deviation ($\pm0.75$), while the second highest is Edinburgh ($\pm0.64$) and the third highest is TBD ($\pm0.52$). This shows that our dataset contains great variation in pedestrian motions. Among the sub-datasets, in CMU and Keio, pedestrians have a low average speed due to a large presence of static pedestrians. In Miraikan, pedestrians' average speed is also low, because the pedestrians' overall walking speed tends to be slow to enjoy the museum, and there is also a moderate amount of static pedestrians (e.g., stopping to examine exhibitions). NUS ($\pm0.90$) and UBonn ($\pm0.87$) have a high variation in average speed.

\subsubsection{Minimum Distance Between People} As shown in Figure~\ref{fig:bev_traj_stat}c, our dataset is smaller than Edinburgh but larger than others. However, our dataset has the largest standard deviation ($\pm4.76$), while the second largest is Edinburgh ($\pm3.5$) and the third largest is TH\"OR ($\pm1.6$). This shows that our dataset has great variation in crowd density. Among the sub-datasets, in high population density locations such as NUS, Miraikan, and Keio, the average crowd density is high, while in the USA and Germany, crowd density is lower. Moreover, NUS ($\pm3.29$), UMich ($\pm5.69$), and UBonn ($\pm9.10$) have particularly great variation in crowd density.

\subsection{Benchmarking SoTA Human Trajectory Prediction Models}

\begin{table}[!t]
\caption{Performance Benchmark of SoTA trajectory prediction models. The data in each cell is ADE/FDE in meters ($m$).}
\label{tab:bev_benchmark}
\renewcommand{\arraystretch}{1.2}
\resizebox{\columnwidth}{!}{
\begin{tabular}{>{\centering\arraybackslash}m{2cm}|>{\centering\arraybackslash}m{2cm}>{\centering\arraybackslash}m{2cm}>{\centering\arraybackslash}m{2cm}}
\toprule
\textbf{Models} & \textbf{ETH/UCY} & \textbf{\ACRONYM} & \textbf{\ACRONYM*}\\
\midrule

\shortstack{SocialGAN\\{\scriptsize \cite{Gupta1}}}
& 0.34 / 0.71 & 0.40 / 0.80 & 0.75 / 1.50\\

\shortstack{AgentFormer\\{\scriptsize \cite{yuan2021agentformer}}}
& 0.23 / 0.47 & 0.30 / 0.58 & 0.54 / 1.05\\

\shortstack{SGNet\\{\scriptsize \cite{wang2022stepwise}}}
& 0.22 / 0.48 & 0.39 / 0.74 & 0.71 / 1.39\\

\shortstack{TUTR\\{\scriptsize \cite{shi2023tutr}}}
& 0.22 / 0.44 & 0.31 / 0.59 & 0.56 / 1.09\\

\shortstack{MoFlow\\{\scriptsize \cite{fu2025moflow}}}
& 0.21 / 0.41 & 0.31 / 0.59 & 0.54 / 1.06\\

\bottomrule
\end{tabular}
}
\begin{tablenotes}
\small
\item[$*$] \textbf{*} indicates the benchmark results on input trajectories with average speed greater than $0.5m/s$.
\end{tablenotes}
\end{table}

An essential use case of the annotated BEV human trajectories is to train trajectory prediction models. To demonstrate this use case, we benchmarked 6 models that represent state-of-the-art trajectory predictors from different eras: SocialGAN \citep{Gupta1}, AgentFormer \citep{yuan2021agentformer}, SGNet \citep{wang2022stepwise}, TUTR \citep{shi2023tutr}, and MoFlow \citep{fu2025moflow}. Consistent with prior benchmarks, the trajectory annotations were downsampled to 2.5Hz. Then, the models were provided 8 timestamps of the trajectory histories (3.2$s$) and generated predictions for the future 12 timestamps (4.8$s$). The models were evaluated stochastically by sampling 20 trajectory predictions, and we took the minimum average displacement errors (ADE) and final displacement errors (FDE) among the sampled trajectories as the results. We trained the models in a cross-validation fashion on the five sub-datasets of the ETH and UCY datasets. Finally, we evaluated the models both on the combined ETH and UCY datasets and on our entire \ACRONYM dataset. The results from the cross-validations are aggregated together by averaging over the minimum 20 ADEs and the minimum 20 FDEs.

The benchmarking results are shown in Table~\ref{tab:bev_benchmark}. The state-of-the-art models perform progressively better on the ETH and UCY combined dataset, consistent with the findings from their respective papers. The newer models also generally perform better on our \ACRONYM dataset, except for SGNet, which only performs slightly better than SocialGAN, and AgentFormer, whose performance is on par with the most recent Moflow. All models perform worse on our dataset when compared to the results achieved on ETH and UCY, with ADE and FDE performances dropping by $0.1m$ and $0.15m$ on average. 

Our dataset contains a mixture of static and dynamic pedestrians, while most pedestrians in ETH and UCY are only dynamic. Similar to the evaluation protocol in the TBD dataset \citep{wang2024tbd}, we additionally benchmarked the models on dynamic pedestrians only. Because the ADE and FDE of the static pedestrians are near zero, we evaluated the models on the dynamic pedestrian trajectories of our dataset only (average speed $>0.5m/s$), and found that the models perform worse. As shown in Table~\ref{tab:bev_benchmark}, the ADE and FDE performances drop by $0.38m$ and $0.72m$ on average, respectively. These performance drops suggest that our \ACRONYM dataset presents a challenging evaluation setting and captures aspects of human motion that are not sufficiently represented in the small-scale ETH and UCY datasets. This indicates that our dataset can serve as a benchmark for assessing model generalization and robustness beyond existing datasets.

\subsection{Comparison across Datasets}

A core contribution of our dataset is that it was collected at diverse locations from 7 different institutions around the world, allowing it to reflect differences in human behavior under different environmental layouts and cultures. In this section, we dive deeper into the annotated human trajectory data on the metric ground plane to identify behavior differences and whether they are related to the varying contexts. %
Based on the semantic tags assigned to the BEV data and following \cite{francis2023principles}, we focus on three common environmental layouts: open areas, wide corridors, and narrow corridors. 
We next identify which sub-datasets provide substantial trajectory coverage for each environmental layout: UBonn, Keio, NUS, Miraikan, and UMich for open areas; CMU, GMU, NUS, Miraikan, and UMich for wide corridors; and CMU, NUS, and Miraikan for narrow corridors.

\subsubsection{Instantaneous Speed Profiles}

\begin{figure*}
    \centering
    \begin{subfigure}{0.95\textwidth}
        \centering
        \includegraphics[width=\textwidth]{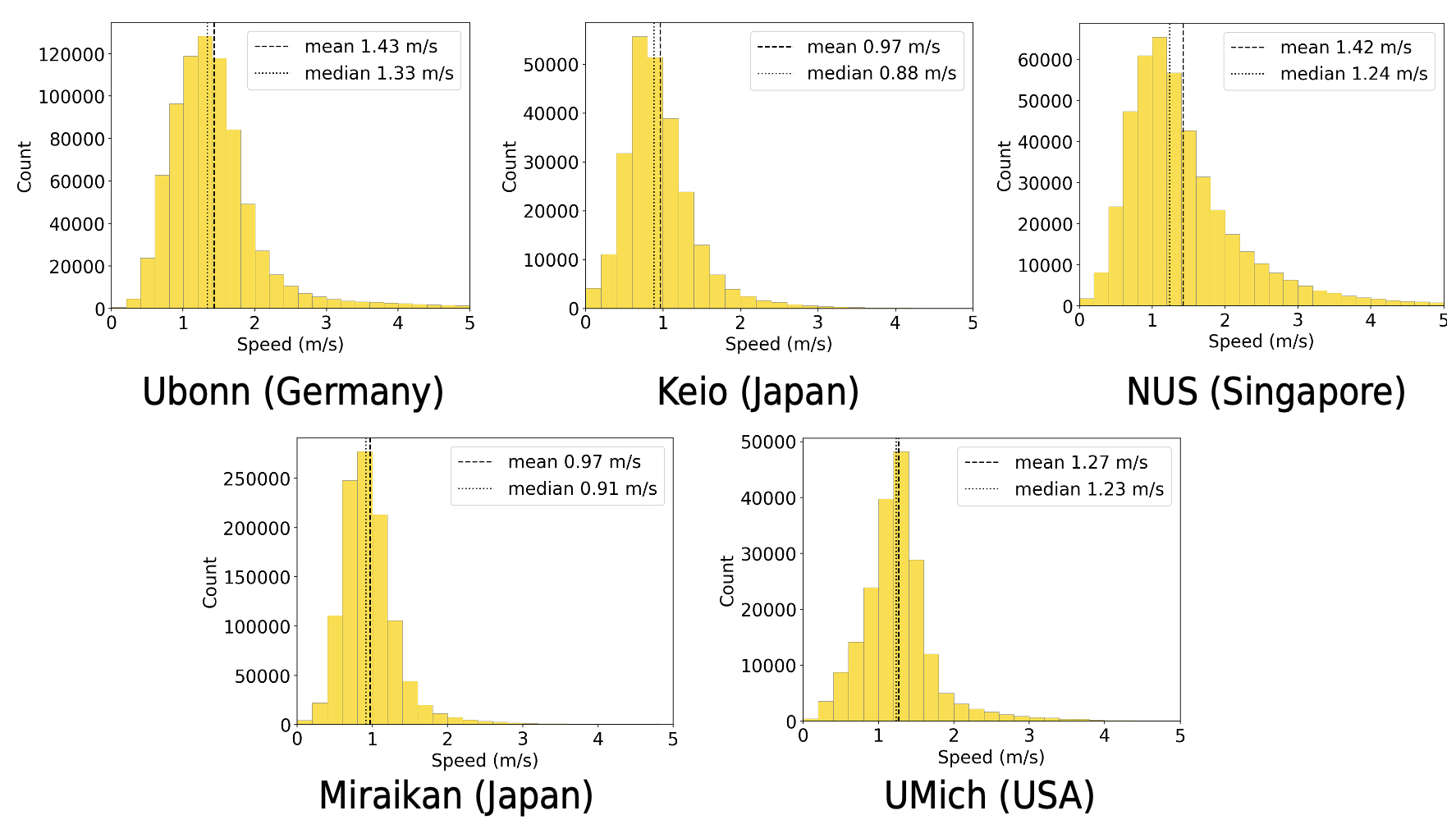}
    \end{subfigure}
    \\
    \begin{subfigure}{0.95\textwidth}
        \centering
        \includegraphics[width=\textwidth]{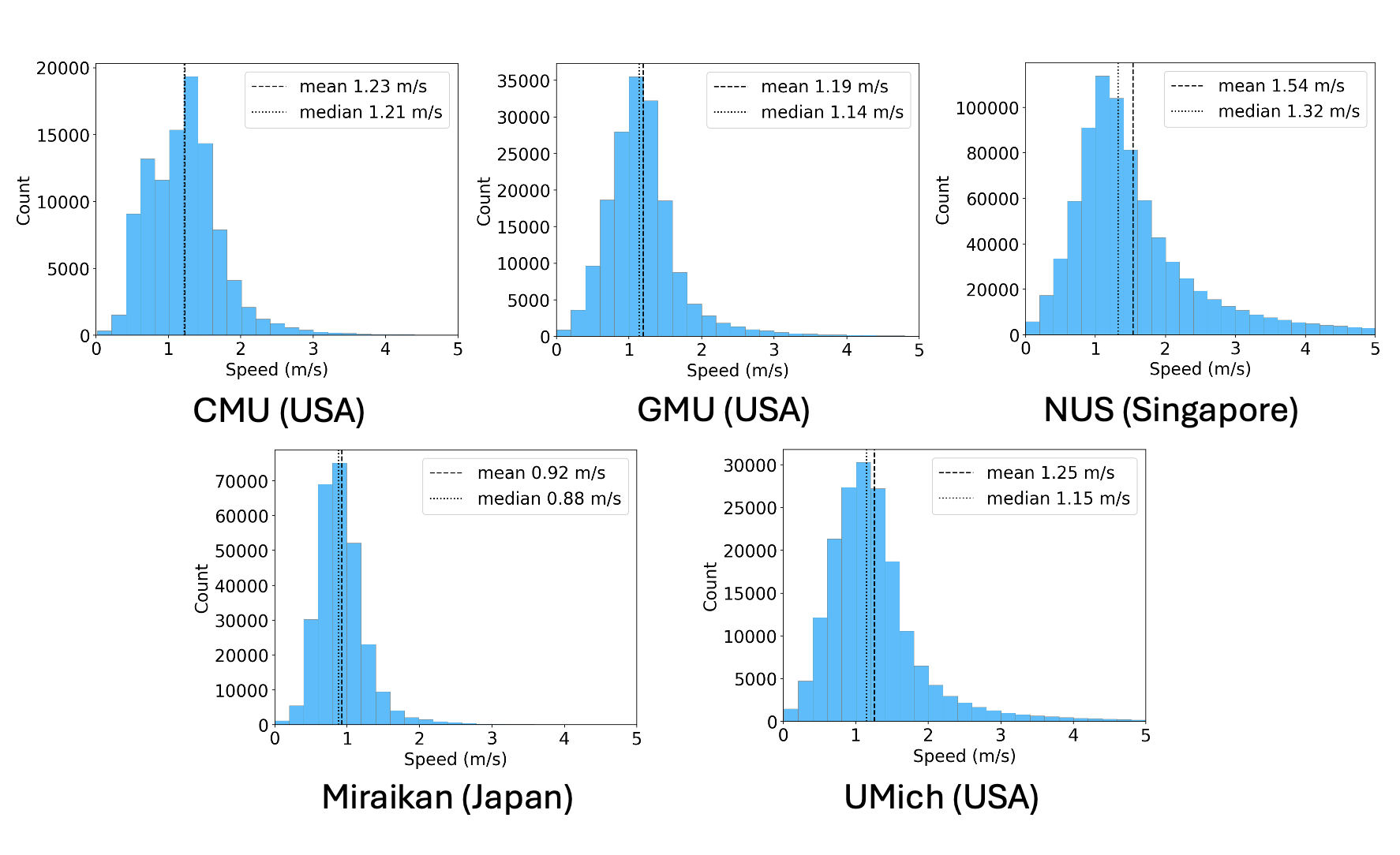}
    \end{subfigure}
    \\
    \begin{subfigure}{0.95\textwidth}
        \centering
        \includegraphics[width=\textwidth]{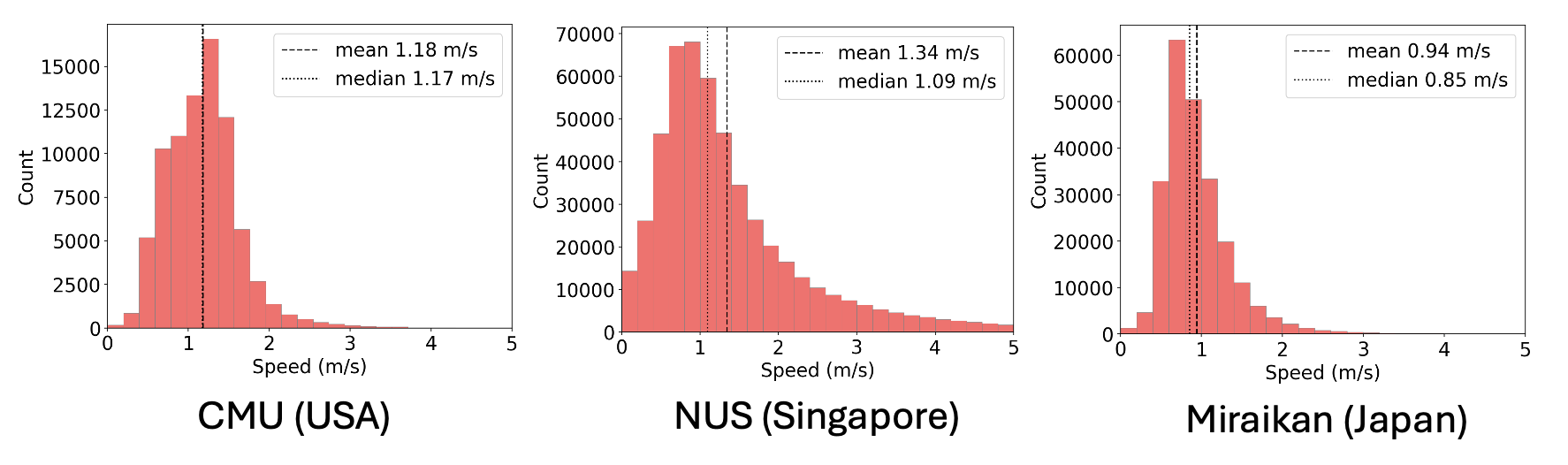}
    \end{subfigure}
    \\
    \caption{Mean, median, and speed distribution of all trajectory segments capped at $5m/s$. Only trajectories with average speed $>0.5 m/s$ are included. The top yellow cluster data are from sub-datasets with significant open area data. The middle blue cluster data are from sub-datasets that contain significant wide corridor data. The bottom red cluster data are from sub-datasets that contain significant narrow corridor data.}
    \label{fig:culture_spd}
\end{figure*}
\begin{figure*}
    \centering
    \begin{subfigure}{0.9\textwidth}
        \centering
        \includegraphics[width=\textwidth]{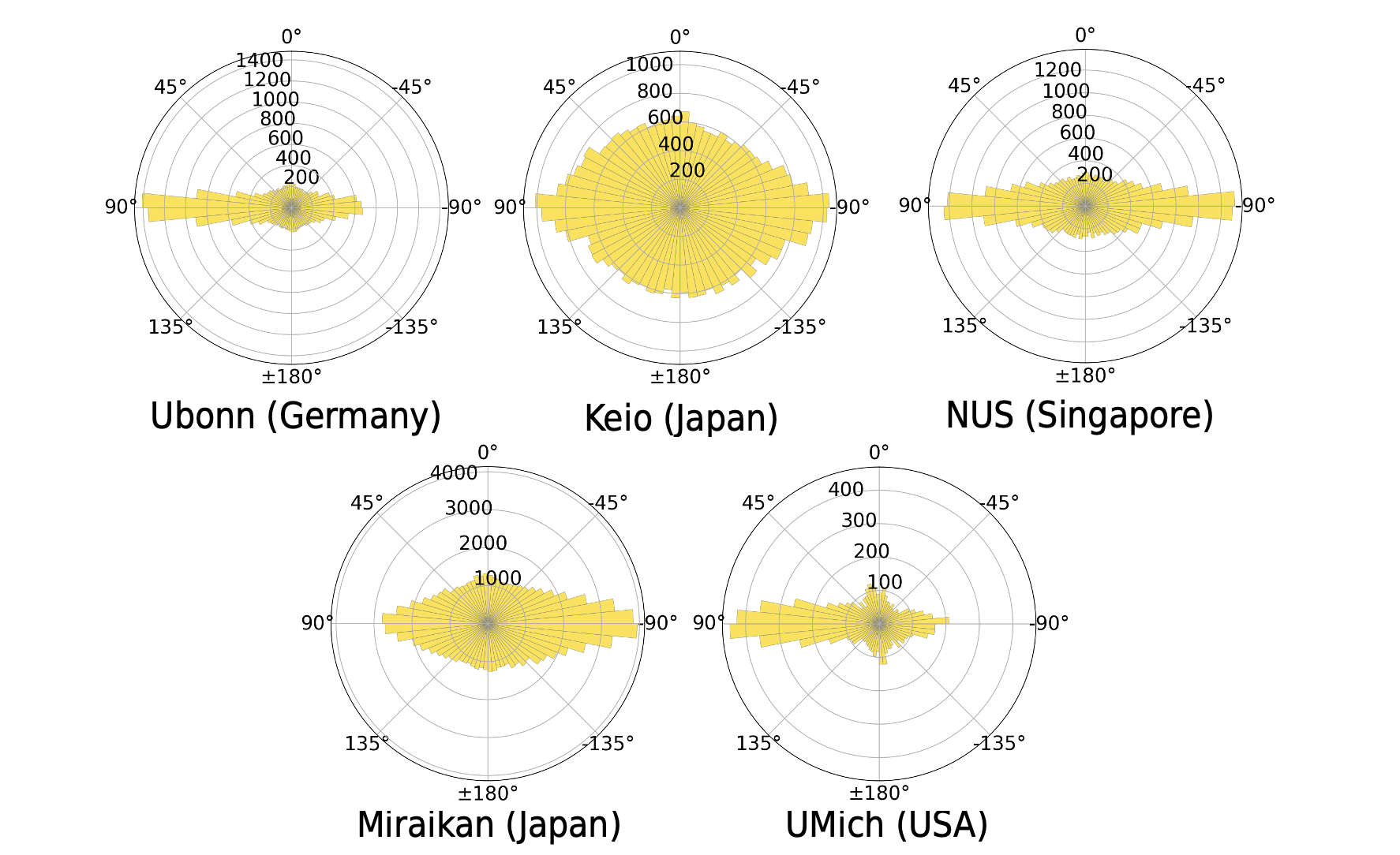}
    \end{subfigure}
    \\
    \begin{subfigure}{0.9\textwidth}
        \centering
        \includegraphics[width=\textwidth]{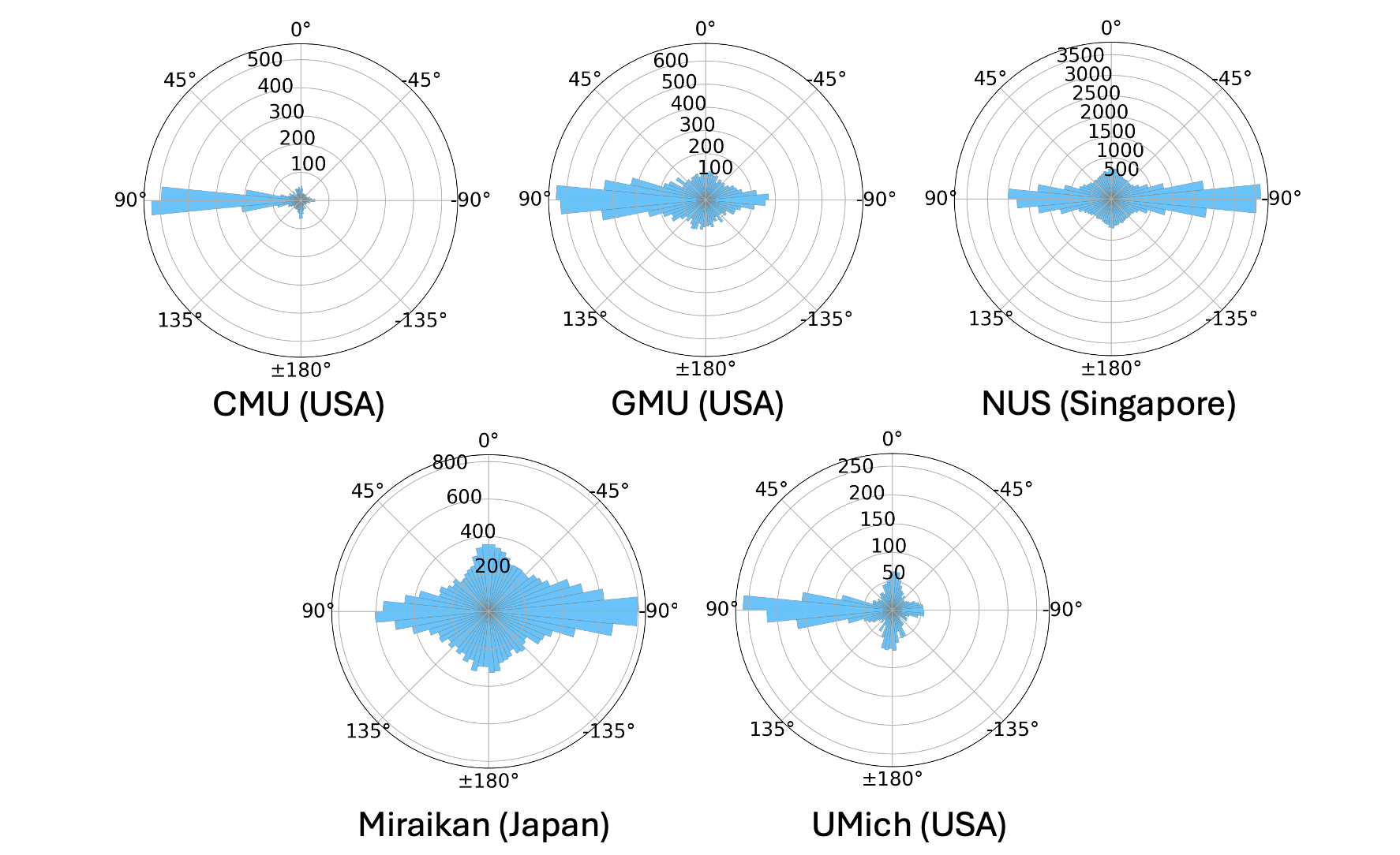}
    \end{subfigure}
    \\
    \begin{subfigure}{0.9\textwidth}
        \centering
        \includegraphics[width=\textwidth]{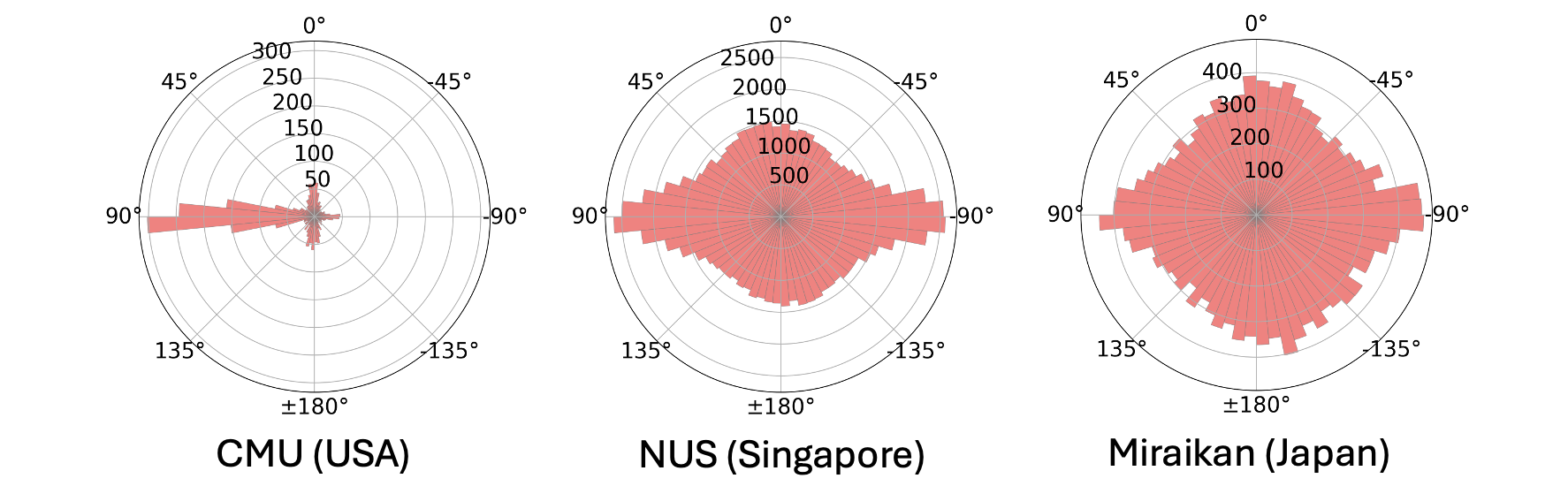}
    \end{subfigure}
    \\
    \caption{Relative direction of pedestrians when interacting with each other. Interaction is defined when two pedestrians get to the closest point to each other, and their distance is $<4m$. The top yellow cluster data are from sub-datasets with significant open area data. The middle blue cluster data are from sub-datasets that contain significant wide corridor data. The bottom red cluster data are from sub-datasets that contain significant narrow corridor data.}
    \label{fig:culture_passing}
\end{figure*}
\begin{figure*}
    \centering
    \begin{subfigure}{0.9\textwidth}
        \centering
        \includegraphics[width=\textwidth]{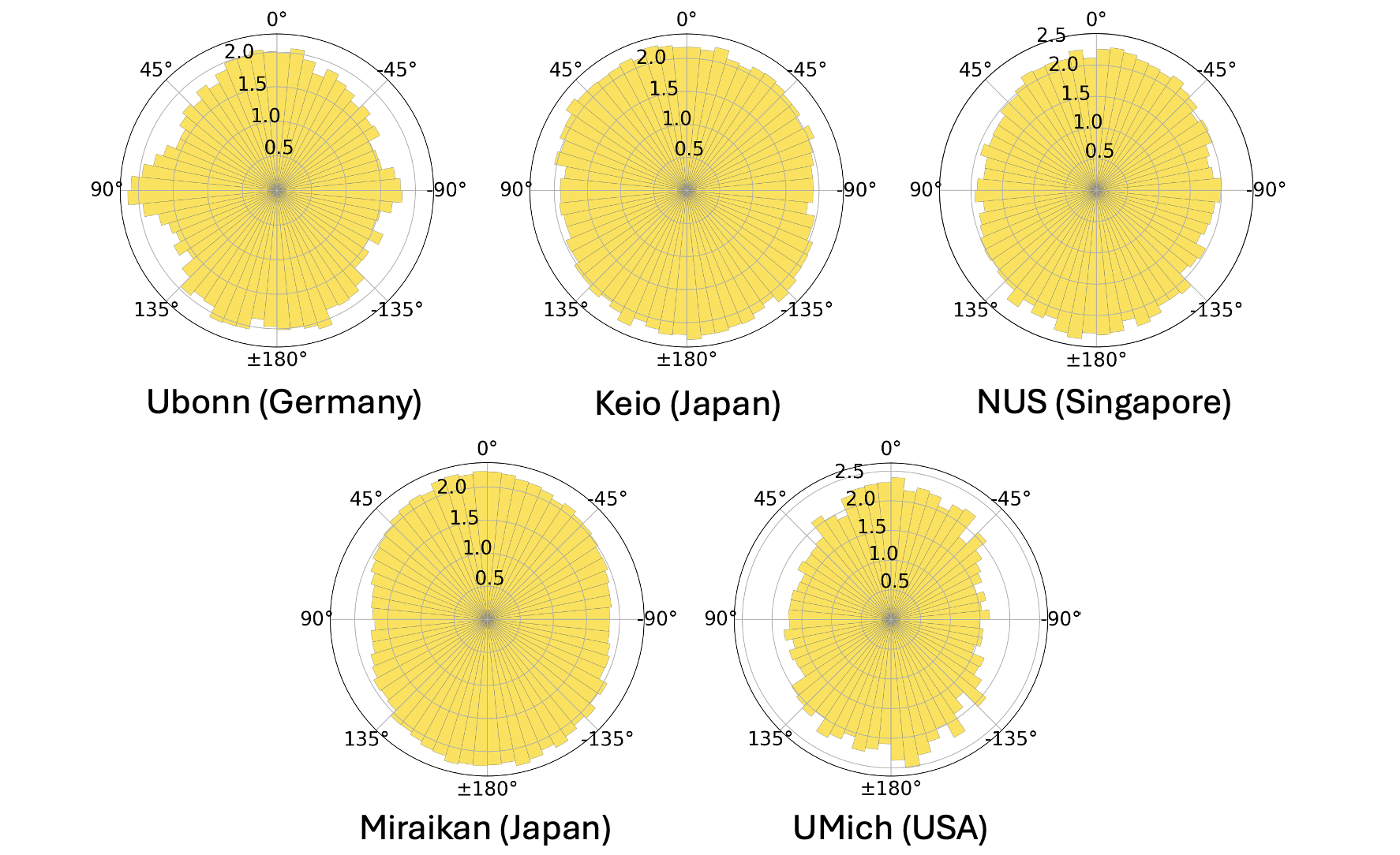}
    \end{subfigure}
    \\
    \begin{subfigure}{0.9\textwidth}
        \centering
        \includegraphics[width=\textwidth]{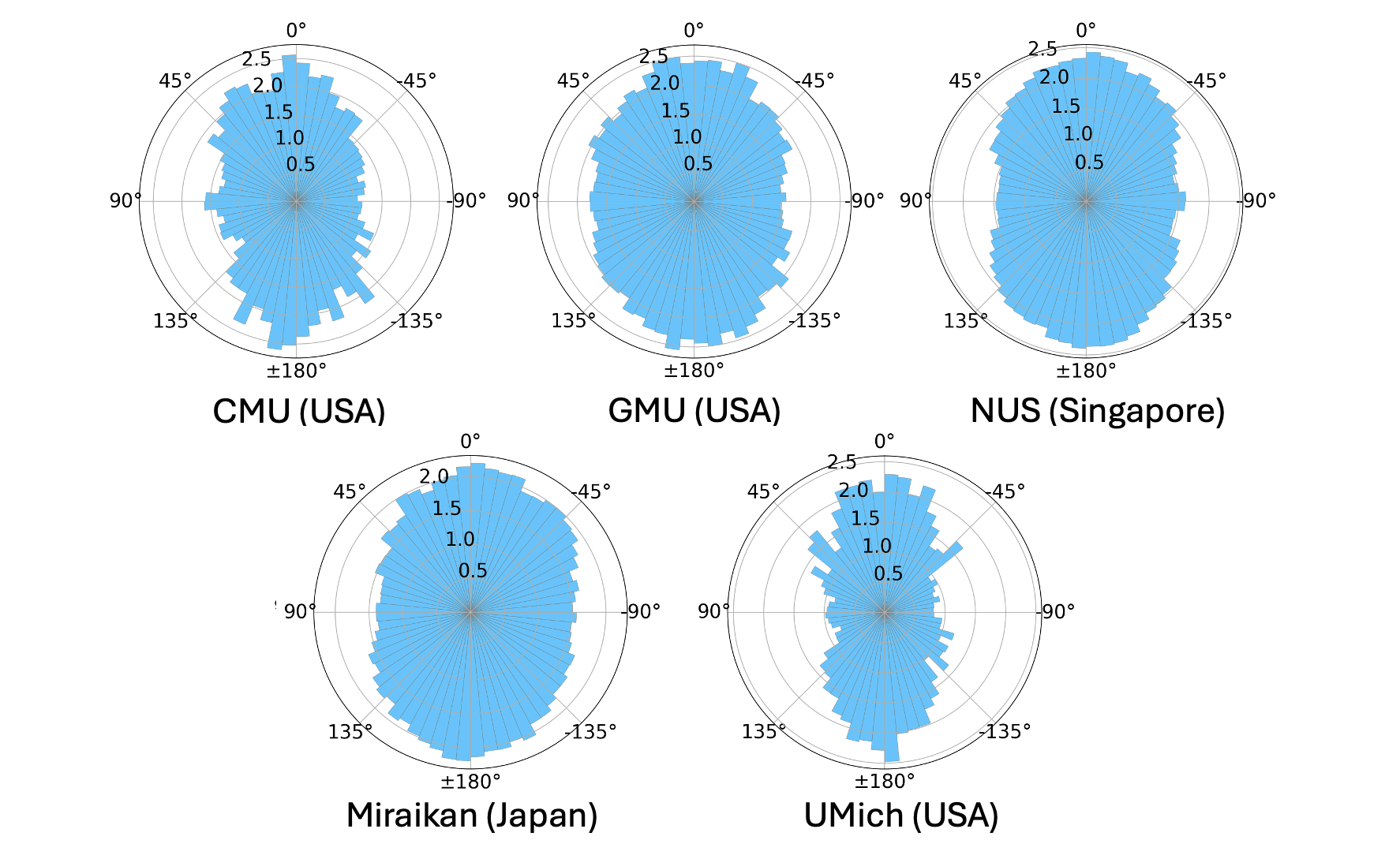}
    \end{subfigure}
    \\
    \begin{subfigure}{0.9\textwidth}
        \centering
        \includegraphics[width=\textwidth]{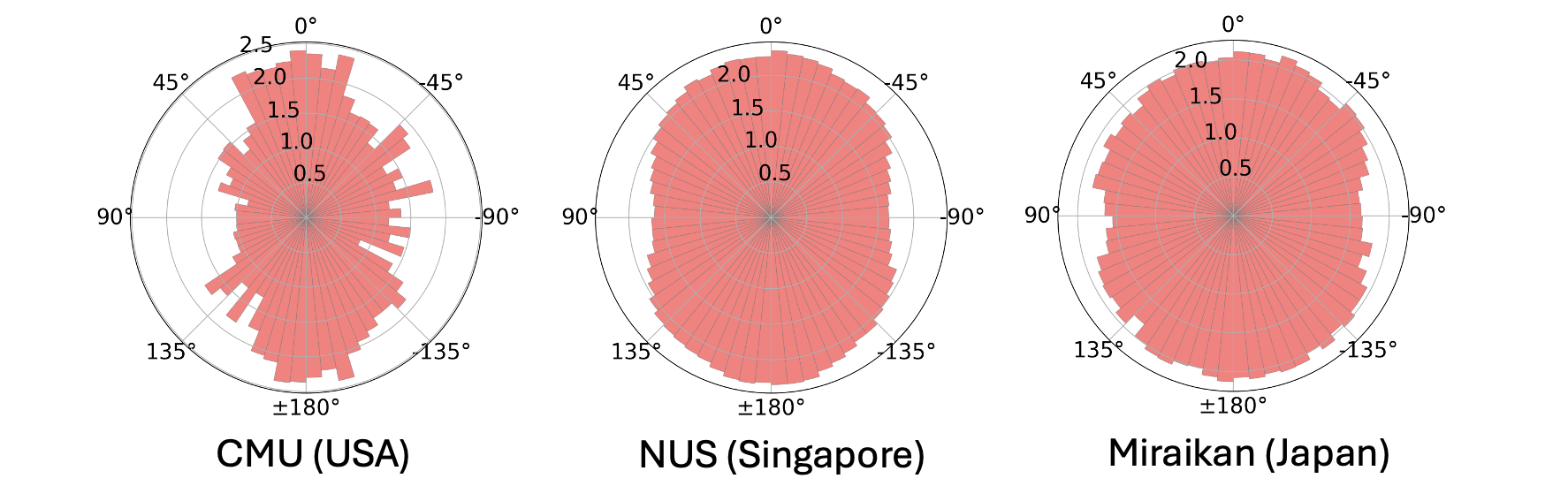}
    \end{subfigure}
    \\
    \caption{Personal space, or average minimum distance of pedestrians when interacting with each other. Interaction is defined when two pedestrians get to the closest point to each other and their distance is $<4m$. The top yellow cluster data are from sub-datasets with significant open area data. The middle blue cluster data are from sub-datasets that contain significant wide corridor data. The bottom red cluster data are from sub-datasets that contain significant narrow corridor data.}
    \label{fig:culture_space}
\end{figure*}
We analyze walking speed at a finer temporal resolution than before. Instead of summarizing each trajectory by its average speed, we use all instantaneous speeds along all the trajectories and group the results by semantic scene tag. To focus on walking behavior, we include only moving pedestrians, defined as trajectories with an average speed greater than $0.5\,m/s$ (some instantaneous speeds may still fall below $0.5\,m/s$). Figure~\ref{fig:culture_spd} shows the resulting speed distributions.

We observe that speed distributions remain similar across different environment layouts. NUS and Miraikan contain data across open areas, wide corridors, and narrow corridors, but the locations' speed distributions are similar. UMich's pedestrian speed distributions are also similar between open areas and wide corridors. CMU's pedestrian speed distributions are also similar between wide corridors and narrow corridors. This shows that culture and general location attributes dictate walking speed more than environmental layout contexts.

Pedestrians in UBonn and NUS exhibit higher walking speeds, which may be partly explained by the prevalence of outdoor settings---all UBonn locations are outdoors, and NUS also includes many outdoor scenes. In contrast, pedestrians in higher-density locations such as Miraikan and Keio tend to walk more slowly, likely due to the limited space available for navigation. Besides, Miraikan is a museum environment, so visitors often slow down or stop completely to view exhibits.

\subsubsection{Passing Side}

Next, we analyze pedestrian behavior in terms of passing side preference. We first identify all pairs of dynamic trajectories that at any point are within $4\,m$ of each other and record the point of closest approach for each pair, which we call an \textit{interaction point}. At each interaction point, we measure the relative direction of one pedestrian with respect to the other's heading. These directions characterize how pedestrians position themselves during passing, overtaking, and following interactions. Figure~\ref{fig:culture_passing} summarizes the results using radial bar charts.

Our analysis suggests that pedestrians' preferred passing side tends to align with the local driving side. UBonn and UMich's open space data show a strong tendency for other interacting pedestrians to approach closest on the left side, meaning people overwhelmingly pass on the right in Germany and in the US. CMU, GMU, and UMich's wide corridor data, as well as CMU's narrow corridor data, all demonstrate the conformity of passing on the right in the US. However, in high-density Asian locations in Japan and Singapore, passing side tendencies become less obvious. While according to Miraikan and NUS's open space and wide corridor data, people tend to pass on the left, this observation is less pronounced in the narrow corridor data from Miraikan and NUS. 

In other observations, in high-density locations such as Keio, Miraikan, and NUS, significantly more interaction points occur at the front and back of pedestrians. In crowded areas, pedestrians tend to follow other pedestrians closely, forming lanes to navigate traffic. Close following can result in more close encounters at the front and the rear.

\subsubsection{Personal Space} Finally, we analyze human behavior in terms of personal space. To perform this analysis, we used the same interaction points as the passing side analysis. 
Rather than counting the number of interactions in each direction, we compute the average closest-approach distance as a function of relative direction. For each direction, we collect all interaction points where one pedestrian is located in that direction relative to another pedestrian's heading and average their Euclidean distances. Repeating this over all directions ($\pm180^\circ$) gives an estimate of direction-dependent personal space.
The resulting aggregated personal spaces by sub-datasets and BEV semantic tags are shown in Figure \ref{fig:culture_space}.

Comparisons between open spaces and corridors in the NUS, Miraikan, and UMich data suggest that lateral personal space becomes narrower in corridor environments. This indicates that environmental layout affects personal space, as pedestrians appear to tolerate smaller side clearances in constrained spaces. However, these lateral differences should not be attributed too strongly to culture, since corridor widths may vary even within the same tagged semantic category (wide or narrow).

In open spaces, frontal personal space appears similar across various locations, with Germany's slightly lower and the US's slightly larger. In wide and narrow corridors, CMU, GMU, UMich, and NUS's data show that personal space tends to be larger at the front in the US and Singapore, closer to $2.5m$.

\section{Conclusion and Discussion}

In this work, we presented \ACRONYM, the first dataset to systematically capture social navigation and pedestrian trajectories across multiple cultural contexts and robot embodiments. \ACRONYM combines contributions from 8 data collection teams from culturally and geographically unique locations, thus providing a unique opportunity to study how navigation behaviors and social norms vary across geography, environment, and robot morphology.

To the best of our knowledge, \ACRONYM is the largest multi-modal social navigation dataset featuring human demonstrations. It provides 29.35 hours of onboard robot data (approximately 1.4 times the size of MuSoHu) and 43.5 hours of pedestrian trajectory data (roughly 7 times the volume of comparable datasets like TBD and TH"OR-MAGNI). Additionally, \ACRONYM introduces unique features, including 5 distinct robot embodiments and explicit robot-crowd interactions through robot speech. To maximize the utility of the dataset, data with no pedestrians is filtered out, tracking information of anonymized pedestrians is included, the dynamic trajectories from top-down BEV videos are human-verified and tagged semantically, and each robot trajectory is tagged with relevant scenarios to ease data lookup. 

Analysis of the dataset showed that \ACRONYM captures more complex scenarios and pedestrian trajectories than prior datasets while also offering interesting insights into pedestrian behaviors across cultures and contexts and embodiment-dependent interaction behaviors. 

In the future, we aim to collect data in more open settings with full maps for better localization, possibly leveraging pedestrian trajectory annotations to enhance localization robustness to dynamic pedestrians. Robust localization in dynamic environments is needed to achieve the projection of pedestrian trajectory annotations onto the robot's egocentric perspective, which unlocks additional utility such as first-person view trajectory prediction \citep{liu2025egotraj}. Currently, only robots used in Miraikan and Keio sub-datasets have accurate localization that is robust to dynamic pedestrians. We additionally aim to expand to multiple robots in the same location to further investigate differences in pedestrian behaviors localized to specific cultures, and train culture and embodiment-aware social navigation policies. 

\section{Ethical Approval}
The collection of the \ACRONYM dataset involved human observational data across multiple international sites. Ethical approval or exemption was obtained by each participating institution's respective review board prior to data collection:
\begin{itemize}
    \item \textbf{Approved via Full/Expedited Review:} Data collection procedures were reviewed and approved by the Institutional Review Boards at Carnegie Mellon University (STUDY2021\_00000199), the University of Extremadura (89//2025 and 214//2025), Keio University, Faculty of Science and Technology (2026-039), and Miraikan's ethical approval was obtained under Keio University and Miraikan's own legal team.
    \item \textbf{Exempt Status:} The review boards at the National University of Singapore (NUS-IRB-2024-476), the University of Michigan (HUM00268385), George Mason University (STUDY00000382), and the University of Bonn determined that the observational data collection in public spaces did not constitute human subjects research requiring full review, granting an exempt status.
\end{itemize}

\section{Funding}

This research project is supported by A*STAR under its National Robotics Programme (NRP) (award M23NBK0053); 
the JST ASPIRE Program (JPMJAP2501);
NSF 2531320 Public Space Robotics: Community-Driven Models for Social Navigation and Communication;
the National Science Foundation (grants 2350352 and 2531320);
the Spanish Government under grant PID2022-137344OB-C31 (MCIN/AEI/10.13039/501100011033/FEDER, UE); and
the German Federal Ministry of Research, Technology and Space (BMFTR) under the Robotics Institute Germany (RIG), grant No. 16ME0999.

\section{Data Accessibility Statement}
The ACME dataset and associated research materials will be hosted at Hugging Face rather than uploaded directly as research data files. Currently, because of privacy concerns, we are sharing the dataset with the reviewers privately via a password-protected Box in the following link: \blackout{REDACTED}. The repository will provide access to multi-modal data, including egocentric RGB, 3D LiDAR, odometry, calibration information, interaction annotations, scenario tags, and context semantic segmentation and human-verified pedestrian trajectory annotations from overhead views.

For the public release, data collected at each location will comply with the privacy, anonymization, and institutional review requirements of the participating data-collection sites. Accordingly, some visual RGB data have been anonymized through face blurring or full-body masking. However, we plan to release the dataset publicly on Hugging Face via a CC BY license. Accompanying the dataset, all annotations tools and processing scripts will also be made publicly available under Apache License 2.0.
\bibliographystyle{SageH}
\bibliography{refs.bib}

\begin{landscape}
\begin{table}[]
   \caption{Comparison with other Social Navigation Datasets}
\begin{tabularx}{\linewidth}{|
  >{\centering\arraybackslash}p{3.0cm}|  %
  >{\centering\arraybackslash}p{1.2cm}|  %
  >{\centering\arraybackslash}p{2cm}|    %
  >{\centering\arraybackslash}X|         %
  >{\centering\arraybackslash}p{2cm}|    %
  >{\centering\arraybackslash}p{1.2cm}|  %
  >{\centering\arraybackslash}p{2cm}|    %
}
\toprule
Dataset       & \# Trajectories & Trajectory Duration (min) & Sensors                                                                                       & Nav Method    & \# Robots & Location (in/out) \\ \midrule
TBD \citep{wang2024tbd}           & 20        & 264                       & 3D LiDAR, RGB-D Camera, 360 camera, IMU                                                         & Human operated        & 1         & Indoors           \\ \midrule
SCAND \citep{karnan2022scand}         & 138           & 522                       & 3D LiDAR, RGB-D Camera, Wheel Odometry, Visual Odometry                                         & Teleop        & 2         & Both              \\ \midrule
MuSoHu \citep{nguyen2023toward}        & 285           & 1138                      & 3D LiDAR, RGB-D Camera, IMU, Visual Odometry, 360 Camera, Microphone                            & Human walking & 0         & Both              \\ \midrule
HuRON \citep{huron}         & 541           & 4500                      & 2D LiDAR, 360 camera, Fisheye camera, Wheel Odometry                                            & Autonomous    & 1         & Indoors           \\ \midrule
L-CAS \citep{lcas}         & 3             & 49                        & 3D LiDAR                                                                                        & Teleop        & 1         & Indoors           \\ \midrule
FLO-BOT \citep{yan2020robot}       & 6             & 27.5                      & 3D LiDAR, RGB-D camera, Stereo Camera, 2D LiDAR, OEM incremental measuring wheel encoder, IMU   & Autonomous    & 1         & Indoors           \\ \midrule
THOR \citep{thor}          & 600           & 60                        & 3D LiDAR, Motion capture system, Eye-tracking Glasses                                           & Autonomous    & 1         & Indoors           \\ \midrule
JRDB \citep{jrdb}          & 54            & 64                        & 3D LiDAR, 2D LiDAR, Omnidirectional Stereo Suite, RGB camera, RGB-D stereo camera, 6D IMU       & Teleop        & 1         & Both              \\ \midrule
Go Stanford 2 \citep{hirose2018gonet} & N/A           & 1002                      & RGB-D Camera, Fisheye camera, Wheel encoders,                                                   & Teleop        & 1         & Both              \\ \midrule
Crowdbot \citep{diego2022crowdbot}      & 110           & 200+                      & 3D LiDAR, RGB-D Camera, Odometry, contact(Force/Torque sensor)                     & Human operated, Autonomous & 1         & Outdoor           \\ \midrule
SIT \citep{bae2023sit}           & N/A           & 20+                       & 3D LiDAR, 5x RGB Cameras, IMU, GPS-RTK                                                          & Teleop        & 1         & Both              \\ \midrule
CityWalker \citep{liu2025citywalker}    & N/A           & 900                       & 3D LiDAR, RGB Cameras, GPS                                                                      & Teleop        & 1         & Outdoor           \\ \midrule
PeRoI \citep{agrawal2026icra}    & 18699           & 67.2                       & 2D trajectory from bird's eye camera                                                                      & Teleop        & 3         & Outdoor           \\ \midrule
Bi$^3$ \citep{stratton2026bi3dataset}    & 185           & 630                       & Motion Capture System, RGB Camera, Depth Camera                                                                      & Autonomous        & 2         & Indoor           \\ \midrule
\textbf{\ACRONYM}          & 3013     &     1761             & 3D-LiDAR, RGB(D) Camera, Overhead BEV Cameras, Odometry                                          & Human operated, Teleop        & 7         & Both             \\ \bottomrule
\end{tabularx}
\label{tab:comparison_socnav}
\end{table}
\end{landscape}

\section{Appendix}
\subsection{Miscellaneous Details}
\subsubsection{Data Curation}
Due to the distributed, multi-site, and multi-embodiment nature of \ACRONYM, the raw data streams differ slightly across subsets in terms of sensor availability, calibration quality, and recovery procedures. While all subsets follow the common data-collection protocols described in the main paper, some recordings experienced hardware or software issues that affected specific modalities. We document these subset-specific notes here to support reproducibility and to help users select the appropriate subset and modalities for downstream tasks.
\begin{itemize}
    \item \textbf{NUS:} To account for sensor misalignment as an effect of robot transport to different locations, we provide session-wise manually corrected calibration values, as corrections on top of the default calibration parameters.

    \item \textbf{Miraikan and Keio:} 2.2 hours of additional data do not contain image streams (due to an image-sensor connection failure). These recordings (which still contain LiDAR and Odometry data) are not included in the total dataset duration reported in the main paper.

    \item \textbf{CMU:} Robot odometry was lost due to a hardware failure. Robot pose was recovered post-hoc using LiDAR-inertial odometry \citep{xu2021fast}. However, velocity information is not available in the released processed data and would require additional post-processing to recover.

    \item \textbf{UEx:} Velocity information was lost due to an issue with the ZED 2i ROS~2 driver. We recovered velocity estimates during post-processing using ICP over dense LiDAR scans.
    
    \item \textbf{UBonn:} A small number of trajectories contain camera motion caused by a mounting issue, which can invalidate the corresponding camera calibration. Another fraction of trajectories contains `engagement' type interaction -- the robot intentionally interacts with pedestrians to elicit reactions (similar to \cite{huron}). These trajectories have been tagged appropriately in the released metadata. 
\end{itemize}
\paragraph{Sensor calibration.}
We additionally provide calibration information for the available onboard sensors in each subset. Calibration procedures varied across platforms depending on the robot hardware, sensor mounting, and information available from the robot or sensor manufacturers. We summarize the subset-specific calibration procedures below.

\begin{itemize}
    \item \textbf{NUS:} We use the default calibration provided by the robot manufacturer for the body-to-LiDAR and body-to-camera transforms. Camera intrinsics were estimated using the standard OpenCV camera calibration pipeline.

    \item \textbf{Miraikan, Keio, CMU:} Camera intrinsics were obtained from the sensor manufacturer. Static extrinsic transforms between the robot body and onboard sensors were set based on physical measurements of the sensor mounting configuration.

    \item \textbf{GMU:} Sensor extrinsics were manually calibrated using the known dimensions of the 3D-printed mounting components.

    \item \textbf{UMich:} The Stretch platform provides a default camera transform through its robot model. For the LiDAR, we used motion capture to register the center of the LiDAR and then computed the transform between the Stretch coordinate frame and the LiDAR coordinate frame in the motion-capture system.

    \item \textbf{UEx:} Sensor extrinsics were obtained from the SolidWorks design of the robot platform. Camera intrinsics were estimated using OpenCV calibration utilities: the standard pinhole camera calibration pipeline for the ZED camera and the fisheye calibration pipeline for the 360-degree camera.

    \item \textbf{UBonn:} Calibration information is provided for the available onboard sensors. As noted above, trajectories in which the camera moved due to the mounting issue have been tagged, since the corresponding camera extrinsics may be invalid for those sequences.
\end{itemize}

\begin{figure}
    \centering
        \includegraphics[width=1\linewidth]{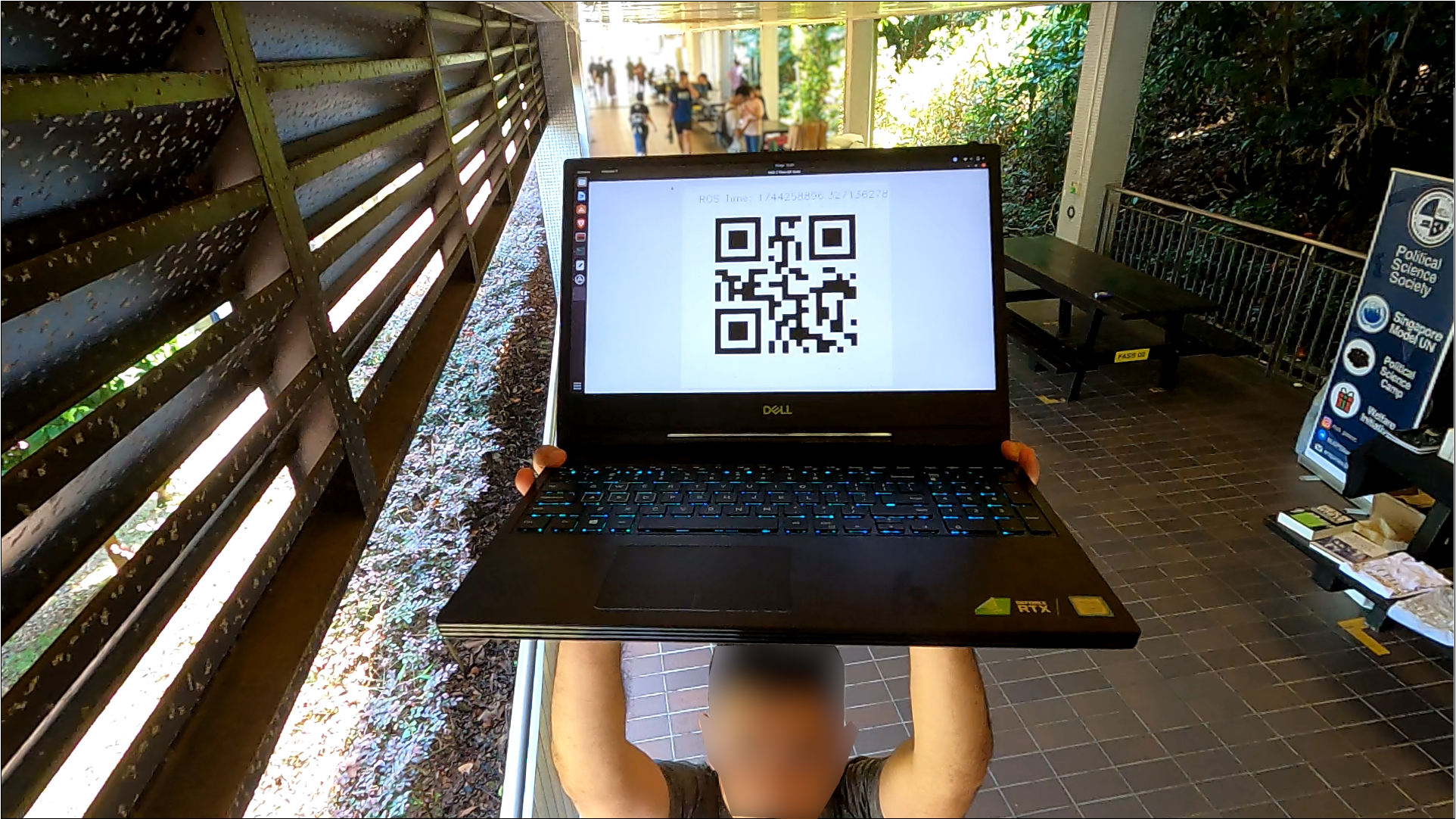}
        \caption{Time Synchronization with timestamp embedded QR Codes}
        \label{fig:time_sync}
\end{figure}

\paragraph{Time-Synchronization between BEV and onboard data}
In order to time-synchronize the BEV camera feed with the on-board robot data, we used a QR-code-based method, which allowed post-hoc temporal syncing of the two data streams. This accommodated the use of commercial BEV cameras with limited SDKs (where direct integration with our ROS-based setup was infeasible). To achieve temporal synchronization across these disconnected systems, the NUS, CMU, U-Mich, GMU, and UBonn teams displayed a dynamic QR code that encoded the current local timestamp to each image sensor at the beginning of a recording session (Fig.  \ref{fig:time_sync}). This permitted alignment of the data streams during post-processing, which facilitated temporal consistency across all image-based sensor streams that were not directly integrated into the ROS network. For Miraikan and Keio, time synchronization was achieved with manual frame selection based on recognizable events (e.g., the frame where someone is about to pick up the tag on the floor). Once a frame with an event/tag was found in both data streams, the rest of the data can be time-synchronized by leveraging ROS bag timestamps for egocentric data and BEV data's stable frame rate to obtain timestamps for any given frame.

\end{document}